\documentclass{article} % For LaTeX2e
\usepackage{iclr2022_conference,times}

\usepackage[utf8]{inputenc} % allow utf-8 input
\usepackage[T1]{fontenc}    % use 8-bit T1 fonts
\usepackage{hyperref}       % hyperlinks
\usepackage{url}            % simple URL typesetting
\usepackage{booktabs}       % professional-quality tables
\usepackage{amsfonts}       % blackboard math symbols
\usepackage{nicefrac}       % compact symbols for 1/2, etc.
\usepackage{microtype}      % microtypography
\usepackage{xcolor}         % colors
\usepackage{algorithm}
\usepackage{algpseudocode}
\usepackage{wrapfig}
\usepackage{graphicx}
\usepackage{subfigure}
\usepackage{booktabs} % for professional tables
\usepackage{macros} % Macros

\usepackage{xcite}
\usepackage{xr}
\makeatletter
\newcommand*{\addFileDependency}[1]{% argument=file name and extension
  \typeout{(#1)}
  \@addtofilelist{#1}
  \IfFileExists{#1}{}{\typeout{No file #1.}}
}
\makeatother

\usepackage[textwidth=2.0cm, textsize=tiny]{todonotes} % for writting

\definecolor{bnfj}{rgb}{0.600000,0.200000,1.000000}

\usepackage{tikz}
\usetikzlibrary{matrix}
\usepackage{pgfplots}
\usepgfplotslibrary{patchplots}
\usetikzlibrary{patterns, positioning, arrows}

\title{Heterogeneous Wasserstein Discrepancy for Incomparable Distributions}

\author{%
  Mokhtar Z.~Alaya \\
  LMAC EA 2222\\
  Université de Technologie de Compiègne\\
  \texttt{elmokhtar.alaya@utc.fr} \\
  \And
  Gilles Gasso \\
  LITIS EA 4108 \\
  INSA-Rouen, Université de Rouen Normandie \\
  \texttt{gilles.gasso@insa-rouen.fr } \\
 \And
  Maxime Bérar \\
  LITIS EA 4108 \\
  Université de Rouen Normandie \\
  \texttt{maxime.berar@univ-rouen.fr} \\
\And
  Alain Rakotomamonjy \\
  Criteo AI Lab \\
  \& LITIS EA 4108, Université de Rouen Normandie\\
  \texttt{alain.rakoto@insa-rouen.fr} \\
}

\iclrfinalcopy % Uncomment for camera-ready version, but NOT for submission.
\begin{document}

\maketitle

\begin{abstract}
Optimal Transport (OT) metrics allow for defining discrepancies between two probability measures. Wasserstein distance is for longer the celebrated OT-distance  frequently-used in the literature, which seeks probability distributions to be supported on the {\it same} metric space. Because  of its high computational complexity, several approximate Wasserstein distances have been proposed based on entropy regularization or on slicing, and one-dimensional Wassserstein computation. In this paper, we propose a novel extension of Wasserstein distance to compare two incomparable distributions, that hinges on the idea of {\it distributional slicing}, embeddings, and on computing the closed-form Wassertein distance between the sliced distributions.  We provide a theoretical analysis of this new divergence, called {\it heterogeneous Wasserstein discrepancy (HWD)}, and we show that it preserves several interesting properties including rotation-invariance.  We show that  the embeddings involved in HWD can be  efficiently learned.  Finally, we provide a large set of experiments illustrating the behavior of HWD as a divergence in the context of generative modeling and in query framework.

\end{abstract}

% \begin{abstract}
% Gromov-Wasserstein (GW) distance is a key tool for manifold learning and cross-domain learning, allowing the comparison of distributions that do not live in the same metric space. Because  of its high computational complexity, several approximate GW distances have been proposed based on entropy regularization or on slicing, and one-dimensional GW computation.  In this paper, we propose a novel approach for comparing two incomparable distributions, that hinges on the idea of {\it distributional slicing}, embeddings, and on computing the closed-form Wassertein distance between the sliced distributions.  We provide a theoretical analysis of this new divergence, called {\it distributional sliced embedding (DSE) discrepancy}, and we show that it preserves several interesting properties of GW distance including rotation-invariance.  We show that  the embeddings involved in DSE can be  efficiently learned.  Finally, we provide a large set of experiments illustrating the behavior of DSE as a divergence in the context of generative modeling and in query framework.
% \end{abstract}

%!TEX root = main.tex

\section{Introduction} % (fold)
\label{sec:introduction}

Optimal Transport-based data analysis has recently found widespread interest in machine learning community, since its significant usefulness to achieve many tasks arising from designing loss functions in supervised learning~\citep{frogner2015nips}, unsupervised learning~\citep{pmlr-v70-arjovsky17a}, %clustering~\citep{ho2017}, 
text classification~\citep{kusnerb2015}, domain adaptation~\citep{courty2017optimal}, generative models~\citep{pmlr-v70-arjovsky17a,salimans2018improving}, computer vision~\citep{bonnel2011,solomon2015} among many more applications~\citep{klouri17,peyre2019COTnowpublisher}. 
Optimal Transport (OT) attempts to match real-world entities through computing distances between distributions, and for that it exploits prior geometric knowledge on the base spaces in which the distributions are valued.  Computing OT distance equals to finding the most cost-efficiency way to transport mass from source distribution to target distribution, and it is  often referred to as the Monge-Kantorovich or Wasserstein distance~\citep{monge1781,kantorovich1942,villani09optimal}.

Matching distributions using Wasserstein distance relies on the assumption that their base spaces must be the same, or that at least a meaningful pairwise distance between the supports of these distributions can be computed. A variant of Wasserstein distance dealing with heterogeneous distributions and overcoming the lack of intrinsic correspondence between their base spaces is Gromov-Wasserstein (GW) distance~\citep{sturm2006,memoli2011GW}. GW distance allows to learn an optimal transport-like plan by measuring how the similarity distances between pairs of supports within each ground space are closed.
It is increasingly finding applications for learning problems in 
shape matching~\citep{memoli2011GW}, graph partitioning and matching~\citep{xu2019neurips}, matching of vocabulary sets between different languages~\citep{alvarezmelis2018gromov}, generative models~\citep{pmlr-v97-bunne19a}, or matching weighted networks~\citep{chowdhury2018}.
Due to the heterogeneity of the distributions, GW distance uses only the relational aspects in each domain, such as the pairwise relationships to compare the two distributions.
As a consequence, the main disadvantage of GW distance is its computational cost as the associated optimization problem is a non-convex quadratic program~\citep{peyre2019COTnowpublisher}, and as few as thousand samples can be computationally challenging. 
% Computing GW distance leads to a nonconvex quadratic program~\citep{peyre2019COTnowpublisher} with an expensive computation for problems with more than a few thousands of points, hence it limits its applications in large-scale settings. 
Based on the approach of regularized OT~\citep{cuturinips13}, in which an entropic penalty is added to the original objective function defining the Wasserstein OT problem, ~\cite{peyre2016Gromov-W} propose an entropic version called {entropic GW discrepancy}, that leads to approximate GW distance. Another approach for scaling up the GW distance is Sliced Gromov-Wasserstein (SGW) discrepancy~\citep{sgw}, which leverages
on random projections on 1D and on a closed-form solution of the 1D-Gromov-Wasserstein.

\begin{figure}[tp]% 12cm 
	~\hfill\includegraphics[width=0.8\textwidth]{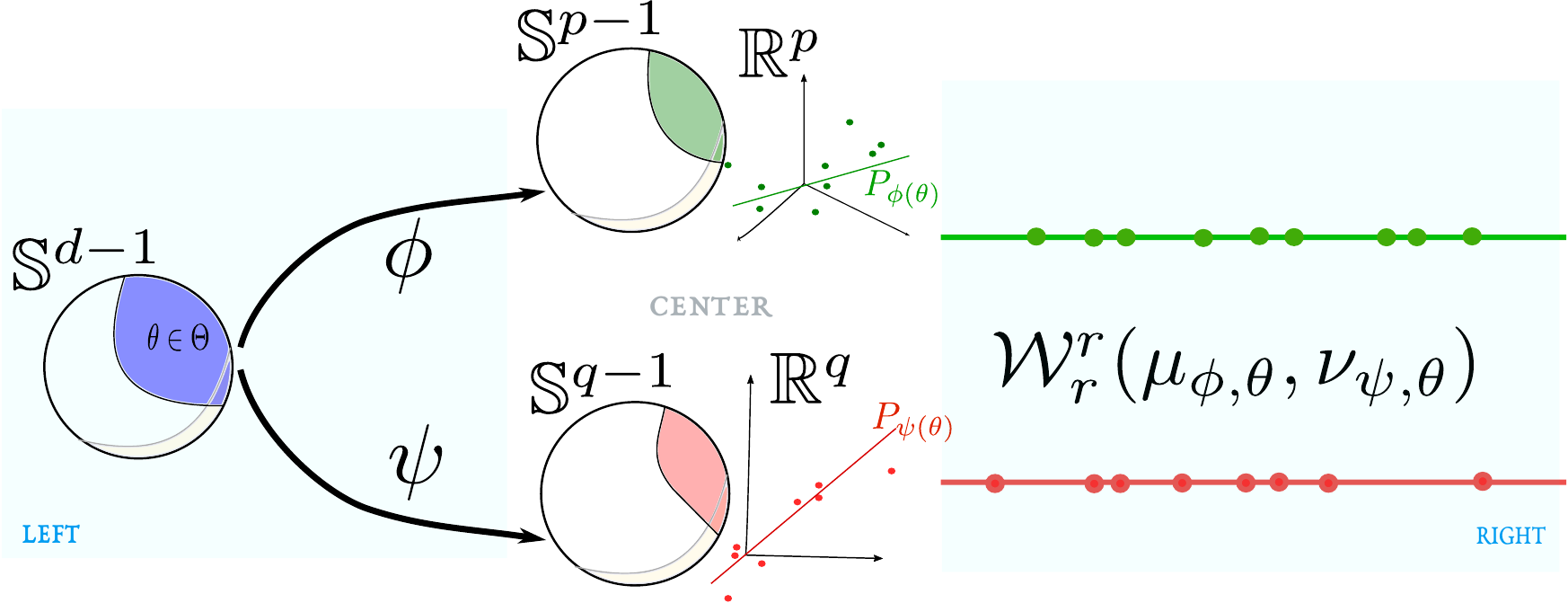}~\hfill
	%\caption{Our approach is based on generating the slicing projections distributions in each of the metric space $\mathbb{R}^p$ and $\mathbb{R}^q$ through the mappings $\phi$ and $\psi$ of a common projection distribution $\Theta$ in $\mathbb{R}^d$. As each of the projected distribution results in a 1D distribution, we can then compute the 1D-Wasserstein distance. It enables us to learn the best projection mapping and to control the distributional part of the generating projection distribution $\Theta$.\label{fig:inform-present}}
	\caption{We measure the discrepancy between two distributions living respectively in  $\mathbb{R}^p$ and $\mathbb{R}^q$.
		Our approach is based on generating random slicing projections distributions in each of the metric spaces $\mathbb{R}^p$ and $\mathbb{R}^q$ through the mappings $\phi$ and $\psi$ of a random projection vector sampled from an optimal distribution $\Theta$ in $\mathbb{R}^d$. As each of the projected distribution results in a 1D distribution, we can then compute 1D-Wasserstein distance. It enables us to learn the best projection mappings $\phi$ and $\psi$  and to optimize over the distributional part of the generating projection distribution $\Theta$.\label{fig:inform-present}}
\end{figure}

In this paper, we take a different approach for measuring the discrepancy
between two heterogeneous distributions. Unlike GW distance that
compares pairwise distances of elements from each distribution, we consider a method that embeds the metric measure
spaces into a one-dimensional space and computes a Wasserstein distance
between the two 1D-projected distributions. The key element of
our approach is to learn two mappings that transform vectors
from the unit-sphere of a latent space to the unit-sphere of the
metric space underlying the two distributions of interest, see Figure~\ref{fig:inform-present}. 
In a nutshell,
we learn to transform a random direction, sampled under an optimal (learned) distribution (optimality being made clear later), from a $d$-dimensional space 
to a random direction into the desired spaces. This approach has
the benefit of avoiding an ad-hoc padding strategy (completion of $0$ of
the smaller dimension distributions to fit the high-dimensional one) as
in SGW method~\citep{sgw}. Another relevant feature of our approach is that
the two resulting 1D distributions are now compared through Wasserstein distance.
This point, in conjunction, with other key aspect of the method, will lead
to a relevant discrepancy between two distributions, called {\it heterogeneous Wasserstein discrepancy (HWD)}. Although we lose
some properties of a distance, we show that HWD is rotation-invariant,
that it is robust enough to be considered as a 
loss for learning generative models between heterogeneous spaces. We also establish that HWD boils down to the recent distributional sliced Wasserstein distance~\citep{nguyen2020distributional} if the two distributions live in the same space and if some mild constraints are imposed on the mappings. 

In summary, our contributions are as follows: 
\begin{itemize}
	\item we propose HWD, a novel slicing-based discrepancy for comparing two distributions living in different spaces. Our chosen formulation is based on comparing 1D random-projected versions of the two distributions using a Wasserstein distance;
	\item The projection operations are materialized by optimally mapping from one common space to the two spaces of interest. We provide a theoretical analysis of the resulting discrepancy and exhibit its relevant properties;
	\item Since the discrepancy involves several mappings that need to be optimized, we depict an alternate optimization algorithm for learning them;  
	\item Numerically, we validate the benefits of HWD in terms of comparison between heterogeneous distributions. We show that it can be used as a loss for generative models or shape objects retrieval with better performance and robustness than SGW on those tasks. 
	
\end{itemize}

\section{Background of OT distances} % (fold)
\label{sec:preliminaries}
For the reader’s convenience, we provide here a brief review of the notations and definitions, that will be frequently used throughout the paper. We start by introducing Wasserstein and Gromov-Wasserstein distances with their sliced versions SW and SGW, where  
% needed to introduce the main results
% the standard notation and
% the definitions that will be frequently used throughout the paper.
% We start here by reviewing basic definitions of the materials needed to introduce the main results, in particular the Wasserstein and Gromov-Wasserstein distances with their sliced versions SW and SGW. 
we consider these distances in the specific case of Euclidean base spaces $(\R^p, \norm{\cdot})$ and $(\R^q, \norm{\cdot})$. We denote $\mathscr{P}(\cX)$ and $\mathscr{P}(\cY)$ the respective sets of probability measures whose supports are contained on compact sets $\cX \subseteq \R^p$ and $\cY \subseteq \R^q$. For $r\geq 1$, we denote $\mathscr{P}_r(\cX)$ the subset of measures in $\mathscr{P}(\cX)$ with finite $r$-th moment $(r \geq 1)$, i.e., $\mathscr{P}_r(\cX) = \big\{\eta \in \mathscr{P}(\cX): \int_{\cX} \norm{x}^r\diff\eta(x) < \infty\big\}.$ %where $\norm{\cdot}$ stands for the Euclidean norm.
For $\mu \in \mathscr{P}(\cX)$ and $\nu \in \mathscr{P}(\cY)$, we write $\Pi(\mu, \nu) \subset \mathscr{P}(\cX \times \cY)$ for the collection of joint probability distributions with marginals $\mu$ and $\nu$, known as couplings, ${\Pi(\mu, \nu) = \big\{\gamma \in \mathscr{P}(\cX\times \cY): \forall A \subset \cX, B \subset \cY, \gamma(A \times \cY) = \mu(A), \gamma(\cX \times B) = \nu(B)\big\}.}$ 
% \begin{align*}
% 	\Pi(&\mu, \nu) = \big\{\gamma \in \mathscr{P}(\cX\times \cY): \forall A \subset \cX, B \subset \cY, \gamma(A \times \cY) = \mu(A), \gamma(\cX \times B) = \nu(B)\big\}.
% \end{align*}

\subsection{OT distances for homogeneous domains}

We here assume that the distributions $\mu$ and $\nu$ lie in the same base space, for instance $p=q$. Taking this into account, we can define the Wasserstein distance and its sliced variant.
\paragraph{Wasserstein distance}
% \begin{definition}% ~\citep{villani09optimal}
% (Wasserstein distance).
The $r$-th Wasserstein distance is defined on $\mathscr{P}_r(\cX)$ by
\begin{equation}
\label{wasserstein-dist}
\mathcal{W}_r(\mu, \nu) = \Big(\inf_{\gamma \in \Pi(\mu, \nu)}\int_{\cX\times \cY}\norm{x - y}^r\diff\gamma(x,y)\Big)^{\frac 1r}.
\end{equation}
% \end{definition}
The quantity $\mathcal{W}_r(\mu, \nu)$ describes the least amount effort to transform one distribution $\mu$ into another one $\nu$. Since the cost distance used between sample supports is the Euclidean one, the infimum in~\eqref{wasserstein-dist} is attained~\citep{villani09optimal}, and any probability $\gamma$ which realizes the minimum is called an {\it optimal transport plan.}
In a finite discrete setting, Problem~\eqref{wasserstein-dist} can be formulated as a linear program, that is challenging to solve algorithmically as its computational cost is of order $\bigO(n^{5/2}\log n)$~\citep{leeSidford2013PathFI}, where $n$ is the number of sample supports. 

Contrastingly, for the 1D case (i.e. $p=1$) of continuous probability measures, the $r$-th Wasserstein distance has a closed-form solution~\citep{rachev1998mass}, namely, $\mathcal{W}_r(\mu, \nu) = (\int_0^1 |F_\mu^{-1}(u) - F_\nu^{-1}(u)|^r \diff t)^{\frac 1r}$
% \begin{equation*}
% \mathcal{W}_r(\mu, \nu) = \bigg(\int_0^1 |F_\mu^{-1}(u) - F_\nu^{-1}(u)|^r \diff t\bigg)^{\frac 1r}
% \end{equation*}
 where $F_\mu^{-1}$ and $F_\nu^{-1}$ are the quantile functions of $\mu$ and $\nu$. For empirical distributions, the 1D-Wasserstein distance is simply calculated by sorting the supports of the distributions on the real line, resulting to a complexity of order $\bigO(n\log n)$. 
This nice computational property motivates the use of sliced-Wasserstein (SW) distance~\citep{rabin-sliced-2011,bonneel-2015-sliced}, where one calculates an (infinity) of 1D-Wasserstein distances between linear projection pushforwards of distributions in question and then computes their average. 

To precisely define SW distance, we consider the following notation. Let $\mbS^{p-1} := \{u \in \R^p: \norm{u} =1\}$ be the unit sphere in $p$ dimension  in $\ell_2$-norm, and for any vector $\theta$ in $\mbS^{p-1}$, we define $P_\theta$ the orthogonal projection onto the real line $\R\theta = \{\alpha \theta: \alpha \in \R\}$, that is $P_\theta(x) = \inr{\theta, x},$ where $\inr{\cdot, \cdot}$ stands for the Euclidean inner-product. Let $\mu_\theta = P_\theta\# \mu$ the measure on the real line called pushforward of $\mu$ by $P_\theta$, that is $\mu_\theta(A) = \mu(P_\theta^{-1}(A))$ for all Borel set $A\subseteq \R.$ We may now define the SW distance.
% \begin{definition}%~\citep{rabin-sliced-2011,bonneel-2015-sliced}
% (Sliced Wasserstein distance).
% Assume that $p=q.$
\paragraph{Sliced Wasserstein distance}
The $r$-th order sliced Wasserstein distance between two probability distributions $\mu, \nu \in \mathscr{P}_r(\cX)$ is given by
\begin{align}% \fint
\label{sliced-W}
\mathcal{SW}_r(\mu, \nu) = \Big(\frac{1}{A_p}\int_{\mbS^{p-1}}\mathcal{W}_r^r(\mu_\theta, \nu_\theta)\diff\theta\Big)^{\frac 1r},
\end{align}% \frac{1}{{\text{surf}}
where $A_p$ is the area of the surface of  $\mbS^{p-1}$, i.e., $A_p = \frac{2\pi^{p/2}}{\Gamma(p/2)}$ with $\Gamma: \R \rightarrow \R$, the Gamma function given as $\Gamma(u) = \int_0^\infty t^{u-1}e^{-t}\diff t.$
% where $\fint_{\mbS^{p-1}} = 1/A_p\int_{\mbS^{p-1}}$
% The last normalized integral can be seen as the expectation for $\theta\sim\sigma^{p-1}$,  the uniform surface measure on $\mbS^{p-1}$, namely,
% $\mathcal{SW}_r(\mu, \nu) = \big(\E_{\theta \sim \sigma^{p-1}}\big[\mathcal{W}_r^r(\mu_\theta, \nu_\theta)\big]\big)^{\frac 1r}.$
% \begin{align*}
% \mathcal{SW}_r(\mu, \nu) = \Big(\E_{\theta \sim \sigma^{p-1}}\big[\mathcal{W}_r^r(\mu_\theta, \nu_\theta)\big]\Big)^{\frac 1r}.
% \end{align*}
% \end{definition}
% \begin{definition}
% (Max-Sliced Wasserstein distance~\citep{Max-SW}).
% Assume that $p=q.$
% The $r$-th order max-sliced Wasserstein distance between two probability distributions $\mu, \nu \in \mathscr{P}_r(\cX)$ is given by
% \begin{equation*}
% \mathcal{MSW}_r(\mu, \nu) = \max_{\theta \in\mbS^{p-1}}	\mathcal{W}_r(\mu_\theta, \nu_\theta),
% \end{equation*}	
% \end{definition}
Thanks to its computational benefits and its valid metric property~\citep{bonnotte:tel-00946781}, the SW distance has recently been used for OT-based deep generative modeling~\citep{kolouri2018sliced,max-SW,wu2019sliced}. Note that the normalized integral in~\eqref{sliced-W} can be seen as the expectation for $\theta\sim\sigma^{p-1}$,  the uniform surface measure on $\mbS^{p-1}$, that is $\mathcal{SW}_r(\mu, \nu) = (\E_{\theta \sim \sigma^{p-1}}[\mathcal{W}_r^r(\mu_\theta, \nu_\theta)])^{\frac 1r}.$ Therefore, the SW distance can be easily approximated via a Monte Carlo sampling scheme by drawing uniform random samples from $\mbS^{p-1}$:  ${\mathcal{SW}}^r_r(\mu, \nu)\approx \frac 1K \sum_{k=1}^K \mathcal{W}_r^r(\mu_{\theta_k}, \nu_{\theta_k})$ where $\theta_1, \ldots, \theta_K \stackrel{i.i.d.}{\sim} \sigma^{p-1}$ and $K$ is the number of random projections.  

\subsection{OT distances for heterogeneous domains}

To get benefit from the advantages of OT in many machine learning applications involving heterogeneous and incomparable domains ($p\neq q$), the Gromov-Wasserstein distance~\citep{memoli2011GW} stands for the basic OT distance dealing with this setting.  % We assume here that $p\neq q$, and more precisely $p<q$ without loss of generality.

\paragraph{Gromov-Wasserstein distance}
% \begin{definition} % ~\citep{memoli2011GW}
% (Gromov-Wasserstein distance).
The $r$-th Gromov-Wasserstein distance between two probability distributions $\mu \in \mathscr{P}_r(\cX)$ and $\nu \in \mathscr{P}_r(\cY)$ is defined by
\begin{align}
\label{gromov-dist}
\mathcal{GW}_r(\mu, \nu) =  \inf_{\gamma \in \Pi(\mu, \nu)}J_r(\gamma)\stackrel{\textrm{def.}}{=}\frac 12 \Big(\iint\limits_{\cX^2\times \cY^2}|\norm{x - x'} -\norm{y- y'}|^r\diff\gamma(x,y) \diff\gamma(x',y')\Big)^{\frac 1r}. 
\end{align}
% where
% \begin{align*}
% J_r(\mu, \nu)= \frac 12 \Big(\iint\limits_{\cX^2\times \cY^2}|\norm{x - x'} -\norm{y- y'}|^r\diff\gamma(x,y) \diff\gamma(x',y')\Big)^{\frac 1r}.	
% \end{align*}
% \end{definition}
Note that $\mathcal{GW}_r(\mu, \nu)$ is a valid metric endowing the collection of all isomorphism classes metric  measure spaces  of $\mathscr{P}_r(\cX) \times \mathscr{P}_r(\cY)$, see Theorem 5 in \citep{memoli2011GW}. The GW distance learns an optimal transport-like plan which transports samples from a source metric space $\cX$ into a target metric space $\cY$, by measuring how the similarity distances between pairs of samples within each space are close. 
Furthermore, GW distance enjoys several geometric properties, particularly translation and rotation invariance. However, its  
major bottleneck consists in an expensive computational cost, since problem~\eqref{gromov-dist} is non-convex and quadratic. A remedy to such a heavy computational burden lies in an entropic regularized GW discrepancy~\citep{peyre2016Gromov-W}, using Sinkhorn iterations algorithm~\citep{cuturinips13}. This latter needs a large regularization parameter to guarantee a fast computation, which, unfortunately, entails a poor approximation of the true GW distance value.
% In the spirit of SW distance, the SGW distance seems 
Another approach to scale up the computation of GW distance is sliced-GW discrepancy~\citep{sgw}. The definition of SGW shows 1D-GW distances between projected pushforward of an artifact zero padding of $\mu$ or $\nu$ distribution. We detail this representation in the following paragraph.

% with respect  of the distributions $\mu$ and $\nu$.

% \begin{definition}% ~\citep{sgw}
% (Sliced Gromov-Wasserstein discrepancy).
\paragraph{Sliced Gromov-Wasserstein discrepancy}
Assume that $p<q$ and let $\Delta$ be an artifact zero padding from $\cX$ onto $\cY$, i.e. $\Delta(x) = (x_1, \ldots, x_p, 0, \ldots, 0) \in \R^q.$
The $r$-th order sliced Gromov-Wasserstein discrepancy between two probability distributions $\mu \in \mathscr{P}_r(\cX)$ and $\nu \in \mathscr{P}_r(\cY)$ is given by
% \begin{align*}
% \mathcal{SGW}_r(\mu, \nu) = \Big(\fint_{\mbS^{p-1}}\mathcal{GW}_r^r((\Delta_{\#}\mu)_\theta, \nu_\theta)\diff\theta\Big)^{\frac 1r},
% \end{align*}
% equivalently
\begin{align}
\mathcal{SGW}_{\Delta, r}(\mu, \nu) = \Big(\E_{\theta \sim \sigma^{q-1}}\big[\mathcal{GW}_r^r((\Delta{\#}\mu)_\theta, \nu_\theta)\big]\Big)^{\frac 1r}.
\label{eq:general_sgw}
\end{align}
% where $\Delta$ is the ``uplifting'' operator which pads each point of the measure with zeros~\citep{sgw}.
% \end{definition}
It is worthy to note that $\mathcal{SGW}_{\Delta, r}$ is depending on the ad-hoc operator $\Delta$, hence the rotation invariance is lost. ~\cite{sgw} propose a variant of SGW that does not depend on the choice of $\Delta$, called Rotation Invariant SGW (RI-SGW) for $p=q$, defined as the minimizer of $\mathcal{SGW}_{\Delta, r}$ over the Stiefel manifold, see ~\citep[Equation 6]{sgw}. In this work, we are interested in calculating an OT-based discrepancy between distributions over distinct domains using the slicing technique. Our approach is different from the SGW one in many points, specifically (and most importantly) we use a 1D-Wasserstein distance between the projected pushforward distributions and not a 1D-GW distance. In the next section, we detail the setup of our approach. 

% : $(i)$ we use a latent space of dimension $d$ to generate two mappings from the unit sphere $\mbS^{d-1}$ to the unit spheres $S$ we use a 1D-Wasserstein distance between the projected pushforward distributions and not a GW one, $(ii)$  the ad-hoc padding operator is 

% section preliminiaries (end)

%!TEX root = main.tex

\section{Heterogeneous Wasserstein discrepancy}
% Distributional sliced embedding OT discrepancy} % (fold)
\label{sec:distributional_sliced_sub_embedding}

Despite the computational benefit of sliced-OT variant discrepancies, they have an unavoidable bottleneck corresponding to an intractable computation of the expectation with respect to uniform distribution of projections. Furthermore, the Monte Carlo sampling scheme can often generate an overwhelming number of irrelevant directions; hence, the larger number of sample projections, the more accurate approximation of sliced-OT values. Recently,~\cite{nguyen2020distributional} have proposed the {\it distributional}-SW distance allowing to find an optimal distribution over an expansion area of informative directions. This  performs the projection efficiently by choosing an optimal number of important random projections needed to capture the structure of distributions.
Our approach for comparing distributions in heterogeneous domains follows a distributional slicing technique combined with OT metric measure embedding~\citep{alaya2020theoretical}.

Let us first introduce additional notations. 
Fix $d \geq 1$ and consider two {\it nonlinear} mappings $\phi: \mbS^{d-1} \rightarrow \mbS^{p-1}$ and $\psi: \mbS^{d-1} \rightarrow \mbS^{q-1}.$ For any constants $C_\phi, C_\psi >0$, we define the following probability measure sets: $\mathscr{M}_{C_\phi} =\big\{\Theta \in \mathscr{P}(\mbS^{d-1}):
\E_{\theta, \theta' \sim \Theta}[|\inr{\phi(\theta), \phi(\theta')}|] \leq C_\phi\big\}$ and $\mathscr{M}_{C_\psi} =\big\{\Theta \in \mathscr{P}(\mbS^{d-1}): \E_{\theta, \theta' \sim \Theta}[|\inr{\psi(\theta), \psi(\theta')}|] \leq C_\psi\big\}.$
We say that $C_\phi$, $C_\psi$  are {\it $(\phi, \psi)$-admissible} constants if the intersection sets $\mathscr{M}_{C_\phi}\cap \mathscr{M}_{C_\psi}$ is not empty. We hereafter denote $\mu_{\phi, \theta} = P_{\phi(\theta)}\# \mu \text{ and }\nu_{\psi, \theta} = P_{\psi(\theta)}\# \nu$ the pushforwards of $\mu$ and $\nu$ by projections over unit sphere $P_{\phi(\theta)}$ and $P_{\psi(\theta)}$, respectively.

\paragraph{Informal presentation}
While the distributions $\mu$ and $\nu$ are valued in different spaces, $\cX \subset \R^p$ and $\cY \subset \R^q,$ any projected distributions  will live in real line, enabling the computation of 1D-Wasserstein distance (Figure ~\ref{fig:inform-present}, right).
In order to generate random 1D projections in each of the spaces, we map a common random projection distribution from $\mbS^{d-1}$ into each of the projection spaces $\mbS^{p-1}$ and $\mbS^{q-1},$  through the mappings $\phi$ and $\psi$ (see Figure~\ref{fig:inform-present}, left). Hence, the main components of the {\it heterogeneous Wasserstein discrepancy} will be the distribution $\Theta \in \mathscr{P}(\mbS^{d-1})$, 
% a subset of the distribution of the unit directions of $\mbS^{d-1}$, 
and the two embeddings $\phi$ and $\psi$ which will be wisely chosen. The resulting directions  $\phi(\theta)$ and $\psi(\theta)$ form the projections $P_{\phi(\theta)}$ and $P_{\psi(\theta)}$ (see Figure~\ref{fig:inform-present}, center) used to compute several 1D-Wasserstein distances.
%\begin{figure}[tp]
%	~\hfill\includegraphics[width=12cm]{./figure/dessin_v2.pdf}~\hfill
%	\caption{Our approach is based on generating the slicing projections distributions in each of the metric space $\mathbb{R}^p$ and $\mathbb{R}^q$ through the mappings $\phi$ and $\psi$ of a common projection distribution $\Theta$ in $\mathbb{R}^d$. As each of the projected distribution results in a 1D distribution, we can then compute the 1D-Wasserstein distance. It enables us to learn the best projection mapping and to control the distributional part of the generating projection distribution $\Theta$.	\label{fig:inform-present}}
%\end{figure}

\subsection{Definition and properties}

Herein we state the formulation of the proposed discrepancy and exhibit its main theoretical properties. 

\begin{definition} 
% Given two probability measures $\mu \in \mathscr{P}_r(\cX)$ and $\nu \in \mathscr{P}_r(\cY)$. 
The heterogeneous Wasserstein discrepancy $(\mathcal{HWD})$ of order $r\geq 1$ between $\mu \in \mathscr{P}_r(\cX)$ and $\nu \in \mathscr{P}_r(\cY)$ reads as
\begin{align} \label{eq:dsse}
\mathcal{HWD}&_r(\mu, \nu)= \inf_{\phi, \psi}\sup_{\Theta \in \mathscr{M}_{C_\phi}\cap\mathscr{M}_{C_\psi}}\Big(\E_{\theta \sim \Theta}\big[\mathcal{W}_r^r(\mu_{\phi, \theta}, \nu_{\psi, \theta})\big]\Big)^{\frac 1r}.
\end{align}
% where the supremum is taken with respect to the intersection of probability measure sets $\mathscr{M}_{C_\phi}$ and $\mathscr{M}_{C_\psi}$ with $C_\phi, C_\psi$ a regularizing $(\phi, \psi)$-admissible constants. %, respectively.
\end{definition}
HWD belongs to a family of projected OT works~\citep{patycuturi2019,pmlr-v89-rowland19a,pmlr-v130-lin21a} with a particularity for seeking nonlinear projections minimizing a sliced-OT variant. 
%A recent work in the same spirit of nonlinear projected OT can be found in~\cite{alaya2020theoretical} where the authors propose to embed the distributions in a common latent space without slicing. 
HWD further inherits the distributional slicing benefit by finding an optimal probability measure $\Theta$ of slices on the unit sphere $\mbS^{d-1}$ coupled with an optimum couple $(\phi, \psi)$ of embeddings. Note that this optimal $\Theta$ verifies  the double conditions $\E_{\theta, \theta' \sim \Theta}[|\cos(\phi(\theta), \phi(\theta'))|] \leq C_\phi$ and $\E_{\theta, \theta' \sim \Theta}[|\cos(\psi(\theta), \psi(\theta'))|] \leq C_\psi$. This gives that $C_\phi, C_\psi \leq 1$, hence the sets $\mathscr{M}_{C_\phi}$ and $\mathscr{M}_{C_\psi}$ belong to $\mathscr{M}_1 = \{\Theta \in \mathscr{P}(\mbS^{d-1})\}$ the set of all probability measures of the unit sphere $\mbS^{d-1}$. It is worthy to note that for small regularizing $(\phi, \psi)$-admissible constants, the measure $\Theta$ is forced to distribute more weights to directions that are far from each other in terms of their angles~\citep{nguyen2020distributional}. 
% \begin{remark}
% Note that the supremum over the probability measure sets $\mathscr{M}_{C_\phi}$ and $\mathscr{M}_{C_\psi}$ guarantees the property $\mathcal{HWD}_r(\mu, \mu) = 0$.
% Indeed, it is essential, since the property $\mathcal{HWD}_r(\mu, \mu) = 0$ will no longer be valid. 
% We mention that we verified empirically this issue. 
% Hence this supremum is of great importance for HWD.
% without taking the supremum over the probability measure sets $\mathscr{M}_{C_\phi}$ and $\mathscr{M}_{C_\psi}$. We verified this issue empirically
% \end{remark}

Now, in order to guarantee the existence of $(\phi, \psi)$-admissible constants, we assume that the couple$(\phi, \psi)$-embeddings are {\it approximately angle preserving}.
\begin{assumption}[Approximately angle preserving property]
\label{ass:ange-preserving}
% Assume that  $\phi$ and $\psi$ are approximately angle preserving embeddings, namely 
For any couple $(\phi, \psi)$-embeddings , 
assume that there exists two non-negative constants $L_\phi$ and $L_\psi,$ such that the following holds
\begin{equation*}
|\inr{\phi(\theta), \phi(\theta')}| \leq L_\phi |\inr{\theta, \theta'}| \text{ and } |\inr{\psi(\theta), \psi(\theta')}| \leq L_\psi |\inr{\theta, \theta'}|, \text{ for all }\theta, \theta' \in \mbS^{d-1}.
\end{equation*}
\end{assumption}
In Proposition~\ref{prop:angle-preserving}, we deliver  lower bounds of the regularizing $(\phi, \psi)$-admissible constants $C_\phi$ and $C_\psi$, depending on the dimension of the latent space $d$ and on the {\it levels} $(L_\phi, L_\psi)$ of approximately angle preserving property. These bounds ensure the non-emptiness of the %probability measure 
sets $\mathscr{M}_{C_\phi}$ and $\mathscr{M}_{C_\psi}$.

\begin{proposition}
\label{prop:angle-preserving}
Let Assumption~\ref{ass:ange-preserving} hold and consider regularizing $(\phi, \psi)$-admissible constants such that $C_\phi \geq \frac{L_\phi\Gamma(d/2)}{\sqrt{\pi} \Gamma((d+1)/2)}$ and $C_\psi \geq \frac{L_\psi\Gamma(d/2)}{\sqrt{\pi} \Gamma((d+1)/2)}$. Then the sets $\mathscr{M}_{C_\phi}$ and $\mathscr{M}_{C_\psi}$ contain the uniform measure $\sigma^{d-1}$ and $\bar{\sigma} = \sum_{k=1}^d \frac 1d \delta_{\theta_k}$, where 
$\{\theta_1, \ldots, \theta_d\}$ forms any orthonormal basis in $\R^d$. Note that by Gautschi's inequality~\citep{gautchi-gamma} for the Gamma function, we have that $C_\phi \geq \frac{L_\phi}{d}$ and $C_\psi \geq \frac{L_\psi}{d}$.
\end{proposition}
Proof of Proposition~\ref{prop:angle-preserving} is presented in Appendix~\ref{sub:proof_of_theorem_th:dsse-distance}. 
Together the admissible constants, the levels of angle preserving property, and the dimension $d$ of the latent space form the hyperparameters set of HWD problem. For settings of large $d$, the admissible constants could take smaller values, that force the measure $\Theta$ to focus on far-angle directions. However, for smaller $d$, we may lose the control on the  distributional part, the set $\mathscr{M}_{C}$ tends to $\mathscr{M}_1$ the entire set of probability measure on $\mbS^{d-1}$, hence it boils down on a standard slicing approach that needs an expensive number of projections to get an accurate approximation. 	
Next, we give a set of interesting theoretical properties characterizing HWD.  
\begin{proposition}
\label{prop:dsse-distance}
For any $r \geq 1$, HWD satisfies the following properties:
\begin{itemize}
\item[(i)] $\mathcal{HWD}_r(\mu, \nu)$ is finite, that is $\mathcal{HWD}_r(\mu, \nu) \leq 2^{\frac{r-1}{r}}(M_r(\mu) + M_r(\nu))$ where $M_r(\cdot)$ is the $r$-th moment of the given  distribution, i.e. $ M_r(\mu) = \big(\int_\cX \norm{x}^r \diff \mu(x) \big)^{\frac 1r}.$
\item[(ii)] $\mathcal{HWD}_r(\mu, \nu)$ is  non-negative, symmetric and verifies $\mathcal{HWD}_r(\mu, \mu) = 0$.

\item[(iii)] % If the regularizing $(\phi, \psi)$-admissible constants verify ${C_\phi \geq \frac{L_\phi}{d}}$ and $C_\psi \geq \frac{L_\psi}{d}$, then one has 
$\mathcal{HWD}_r(\mu, \nu)$ has a discrepancy equivalence given by %  with a (near)-max-sliced Wasserstein distance~\cite{max-SW} % , i.e. ${\big(\frac{1}{d}\big)^{\frac 1r}\inf_{\phi, \psi} \max_{\theta \in \mbS^{d-1}} \mathcal{W}_r(\mu_{\phi, \theta}, \nu_{\psi, \theta}) \leq \mathcal{HWD}_r(\mu, \nu) \leq \inf_{\phi, \psi} \max_{\theta \in \mbS^{d-1}} \mathcal{W}_r(\mu_{\phi, \theta}, \nu_{\psi, \theta})}.$
\begin{equation*}
\big(\frac{1}{d}\big)^{\frac 1r}\inf_{\phi, \psi} \max_{\theta \in \mbS^{d-1}} \mathcal{W}_r(\mu_{\phi, \theta}, \nu_{\psi, \theta}) \leq \mathcal{HWD}_r(\mu, \nu) \leq \inf_{\phi, \psi} \max_{\theta \in \mbS^{d-1}} \mathcal{W}_r(\mu_{\phi, \theta}, \nu_{\psi, \theta}).
\end{equation*}

\item[(iv)] For $p=q$, HWD is upper bounded by the distributional sliced Wasserstein distance. %~\cite{nguyen2020distributional}. %, i.e. $\mathcal{HWD}_r(\mu, \nu) \leq \mathcal{DSW}_r(\mu, \nu)$.

\item[(v)] $\mathcal{HWD}_r$ is rotation invariant, namely, $\mathcal{HWD}_r(R\#\mu, Q\#\nu) = \mathcal{HWD}_r(\mu, \nu)$, for any $R \in \mathcal{O}_p =\{R \in \R^{p\times p}: R^\top R = I_p\}$ and $Q\in \mathcal{O}_q =\{Q \in \R^{q\times q}: Q^\top Q = I_q\}$, the orthogonal group of rotations of order $p$ and $q$, respectively.

\item[(vi)] Let $T_\alpha$ and $T_\beta$ be the translations from $\R^p$ into $\R^p$ and from $\R^q$ into $\R^q$ with vectors $\alpha$ and $\beta$, respectively. Then 
% Let $T_\alpha: \R^p \rightarrow \R^p$, $x \mapsto x+\alpha$ and $T_\beta: \R^q \rightarrow \R^q$, $y \mapsto y + \beta$. Then
% $\mathcal{HWD}_r$ is quasi-translation invariant, i.e., 
${\mathcal{HWD}_r(T_\alpha\#\mu, T_\beta\#\nu)\leq 2^{r-1}\big(\mathcal{HWD}_r(\mu, \nu) + \norm{\alpha}+ \norm{\beta})\big).}$ % \sup_{\Theta \in \mathscr{M}_{C_\phi}\cap\mathscr{M}_{C_\psi}}\big(\Theta(\mbS^{d-1})\big)^{\frac 1r}}.$ 
\end{itemize}
\end{proposition}
Proof of Proposition~\ref{prop:dsse-distance} is given in Apprendix~\ref{sub:proof_of_prop:angle-preserving}. From property $(i)$ HWD is finite provided that the distributions in question have a finite $r$-th moments. 
Note that the supremum over the probability measure sets $\mathscr{M}_{C_\phi}$ and $\mathscr{M}_{C_\psi}$ guarantees the property $\mathcal{HWD}_r(\mu, \mu) = 0$. For $p=q$, if the infimum over the couple $(\phi, \psi)$-embedding  in $(iii)$ is realized in the  identity mappings, then HWD verifies a metric equivalence with respect to the max-sliced Wasserstein distance~\citep{max-SW}.
% the lower bound is exactly $(\frac{1}{d})^{\frac 1r} \max$-$\mathcal{SW}(\mu, \nu)$ (the max-sliced Wasserstein distance~\cite{max-SW}). 
The property $(v)$ highlights a rotation invariance of HWD, which is well verified by the GW distance.

% \subsection{Metric equivalence} % (fold)
% \label{sub:metric_equivalence}

% % subsection metric_equivalence (end)

% subsection computation (end)
% section distributional_sliced_sub_embedding (end)
%!TEX root = main.tex

\subsection{Algorithm	} % (fold)
\label{sub:computation}

% \begin{figure}
% \begin{center}
% 	\includegraphics[width=11cm]{./figure/dessin_implem.pdf}
% 	\end{center}
% 	\caption{.}
% \end{figure}

Computing HWD requires a resolution of 
an optimization  problem as given in~\eqref{eq:dsse}. 
In what follows, we propose an algorithm for computing an approximation of
this discrepancy based on samples $X = \{x_i\}_{i=1}^n$ from $\mu$  and
samples  $Y = \{y_j\}_{j=1}^m$ from $\nu$. 
At first, let us note that we have min-max optimization to solve; the minimization occuring over the embeddings $\phi$ and $\psi$ and the maximization
over the distributions on the unit-sphere. This maximization problem is
challenging due to both the constraints and because we optimize over distributions. Similarly to \cite{nguyen2020distributional}, we approximate
the problem by replacing the constraints with regularization terms and
by replacing the optimization over distributions by an optimization
over a push-forward of the uniform probability measure $\sigma^{d-1}$ by a Borel measurable function $f: \mbS^{d-1} \rightarrow \mbS^{d-1}$.  Hence, assuming that we have drawn from a uniform 
distribution $K$ directions $\{\theta_k\}_{k=1}^K$, the numerical approximation of 
HWD is obtained by solving the following problem:
{\small{  
\begin{align*}\label{eq::first}
\min_{\phi,\psi}  &~~~\max_f \Big\{L_1 \stackrel{\textrm{def.}}{=}\Big(\frac{1}{K}\sum_{k=1}^{K} \mathcal{W}^r_r \big(X^\top \phi [f(\theta_k)],
  Y^\top \psi[f(\theta_k)] \big) \Big )^{1/r}\Big\}   \\
  \notag
   & +  \min_f  \lambda_C \Big\{L_2\stackrel{\textrm{def.}}{=}
  \sum_{k,k^\prime}\phi [f(\theta_k)]^\top \phi [f(\theta_{k^\prime})] 
   + 
  \sum_{k,k^\prime}\psi [f(\theta_k)]^\top \psi [f(\theta_{k^\prime})] \Big\}   \\
  \notag
  & + \min_f \lambda_a \Big \{L_3\stackrel{\textrm{def.}}{=} \sum_{k,,k^\prime} \big(\phi [f(\theta_k)]^\top \phi [f(\theta_{k^\prime})] - \theta_k^\top \theta_{k^\prime} \big)^2 + \sum_{k,k^\prime} \big(\psi [f(\theta_k)]^\top \psi [f(\theta_{k^\prime})] - \theta_k^\top \theta_{k^\prime} \big)^2 \Big\}
\end{align*}
}
}
where the first term in the optimization is related to the sliced Wasserstein, the second term is related to the regularization term associated to $\E_{\theta, \theta' \sim \Theta}[|\inr{\phi(\theta), \phi(\theta')}|],$ and  $\E_{\theta, \theta' \sim \Theta}[|\inr{\psi(\theta), \psi(\theta')}|],$ 
and the third term is the angle-preserving regularization term. Note
that the $\min$ term with respect to $f$ is due to the fact that we want those regularizers
to be small.
Two hyperparameters $\lambda_C$ and $\lambda_a$ control the impact of these two regularization terms.
In practice, $\phi$, $\psi$ and $f$ are parametrized as deep neural networks and the min-max problem is solved by an alternating optimization scheme : (a) optimizing over $f$ with $\psi$ and $\phi$ fixed then (b) optimizing over $\psi$ and $\phi$ with $f$ fixed. 
Some details of the algorithms are provided in Algorithm~\ref{algo:computeDSSE}.

Regarding computational complexity, if we assume that the mapping $f, \phi, \psi$ 
are already trained, that we have $K$ projections, and that $\phi[f(\theta_k)]$, $\psi[f(\theta_k)]$ are precomputed, then
 the computation of HWD (line 25 of Algorithm~\ref{algo:computeDSSE}) is in 
 $O(K(n \log n + np + nq))$, where $n$ is the number of samples
 in $X$ and $Y$. When taking into account the full optimization process, then the complexity depends on the 
 number of times we compute the full objective function we are optimizing. Each evaluation requires the computation of the
 sum in $L_1$ which is  $O(K(n \log n + np + nq))$ and the
 two regularization terms $L_2$ and $L_3$ require both $O(K^2(p+q + 2d))$.
 Note that in terms of computational complexity, SGW is
 $O(Kn\log \ n)$ whereas HWD is $O(T N K n \log n)$, with $T\times N$ being the global number of objective function evaluations.
Hence, complexity is in favor of SGW. However, one should note
that in practice, because we optimize over
the distribution of the random projections, we usually need less slices than SGW and thus depending on the problem, $TNK$ can be of the same magnitude than the number of slices involved in SGW  (similar findings  have been highlighted for Sliced Wasserstein distance \citep{nguyen2020distributional}).

\begin{algorithm}[t]
	\caption{\textsc{Computing Heterogeneous Wasserstein Discrepancy (see \eqref{eq:dsse})}}
	\label{algo:computeDSSE}
	\begin{algorithmic}[1]
		\State {\bfseries Input:} Source and target samples: $(X, \mu)$ and $(Y, \nu);$ order $r$; the set of random direction $\{\theta_k\}_{k=1}^K$; $T$ number of global iterations; $N$ number of iterations for each of the alternate scheme;
		\State {\bfseries Output:} HWD
		\Function{Compute Losses}{$X$, $Y$, $\phi [f(\theta_k)]$ , $\psi [f(\theta_k)]$}
		 % \State {\bfseries Input:} Epochs $(E)$, batch sizes $(B);$
			\State compute the average of % closed-form 
			1D-Wasserstein $L_1$ between $X^\top \phi [f(\theta_k)]$ and $Y^\top \psi [f(\theta_k)]$
			%\State compute the average of the closed-form 1D Wasserstein $L_1$ between $X^\top \phi [f(\theta_k)]$ and $Y^\top \psi [f(\theta_k)]$
			\State compute the inner-product penalty $L_2$
			\State compute the angle-preserving penalty $L_3$

		\EndFunction

		\For{$t=1,\cdots, T$}% {\texttt{\\ $T$ is the global number of iterations}}
		\State fix $\phi$ and $\psi$
		\For{$i=1,\cdots,N$}
			%\For{$k=1,\cdots,K$}
			\State compute $\phi [f(\theta_k)]$ and $\psi [f(\theta_k)]$
			\State $L_1, L_2, L_3 \leftarrow $  Compute Losses ($X$, $Y$, $\phi [f(\theta_k)]$ , $\psi [f(\theta_k)]$))
		    \State $L = - L_1$ + $L_2$ + $L_3$
		    \State $f \leftarrow f - \gamma_k \nabla L$
			%\EndFor
		\EndFor
		\State fix $f$
		\For{$i=1,\cdots,N$}
			%\For{$k=1,\cdots,K$}
			\State compute $\phi [f(\theta_k)]$ and $\psi [f(\theta_k)]$
\State $L_1, L_2, L_3 \leftarrow $  Compute Losses ({$X, Y$, $\phi [f(\theta_k)]$ , $\psi [f(\theta_k)]$})
		\State $L = L_1$ + $L_2$ + $L_3$
		\State $\phi \leftarrow \phi - \gamma_k \nabla L$
		\State $\psi \leftarrow \psi - \gamma_k \nabla L$
		%\EndFor
		\EndFor
		\EndFor
		\State HWD $\leftarrow$ compute the average over $\{\theta_k\}_{k=1}^K$ of closed-form 1D Wasserstein  between $X^\top \phi [f(\theta_k)]$ and $Y^\top \psi [f(\theta_k)]$
		\State {\bfseries Return:} {HWD}
	\end{algorithmic}
\end{algorithm}

% subsection computation (end)
% section distributional_sliced_sub_embedding (end)
%!TEX root = main.tex

\section{Numerical experiments} % (fold)
\label{sec:numerical_experiments}
% In this section, we want to analyze the HWD discrepancy and to 
% exhibit its rotation-invariant property. We will compare its performance
% with SGW in a generative model context.
In this section, we analyze HWD,  exhibit its rotation-invariant property, and compare its performance with SGW in a generative model context.

\paragraph{Translation and Rotation}

\begin{figure}%[htbp] % width=0.16
\begin{minipage}[b]{0.46\linewidth}
~\hfill	\includegraphics[width=0.5\textwidth]{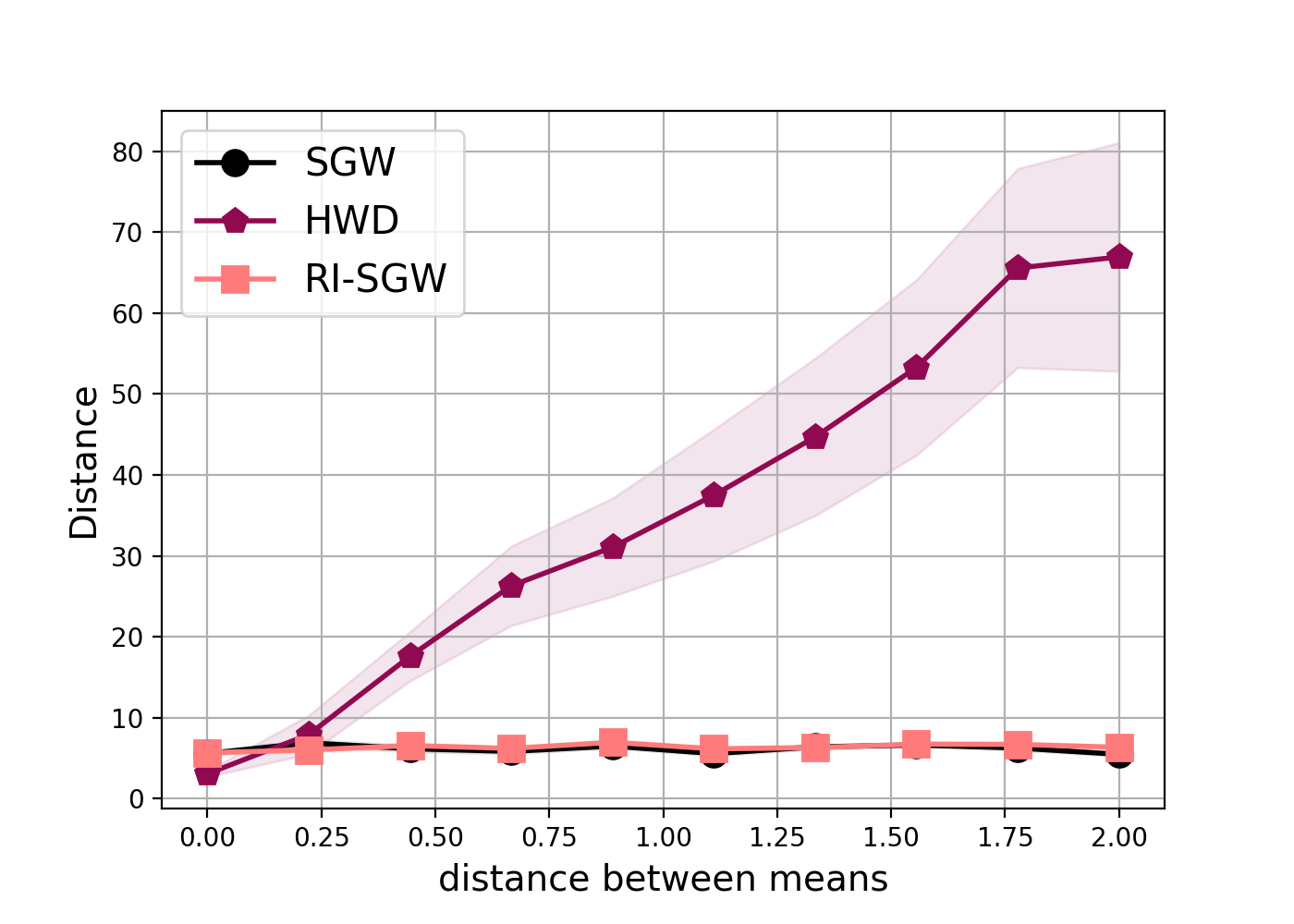}~\hfill~ % width=3cm
	\includegraphics[width=0.5\textwidth]{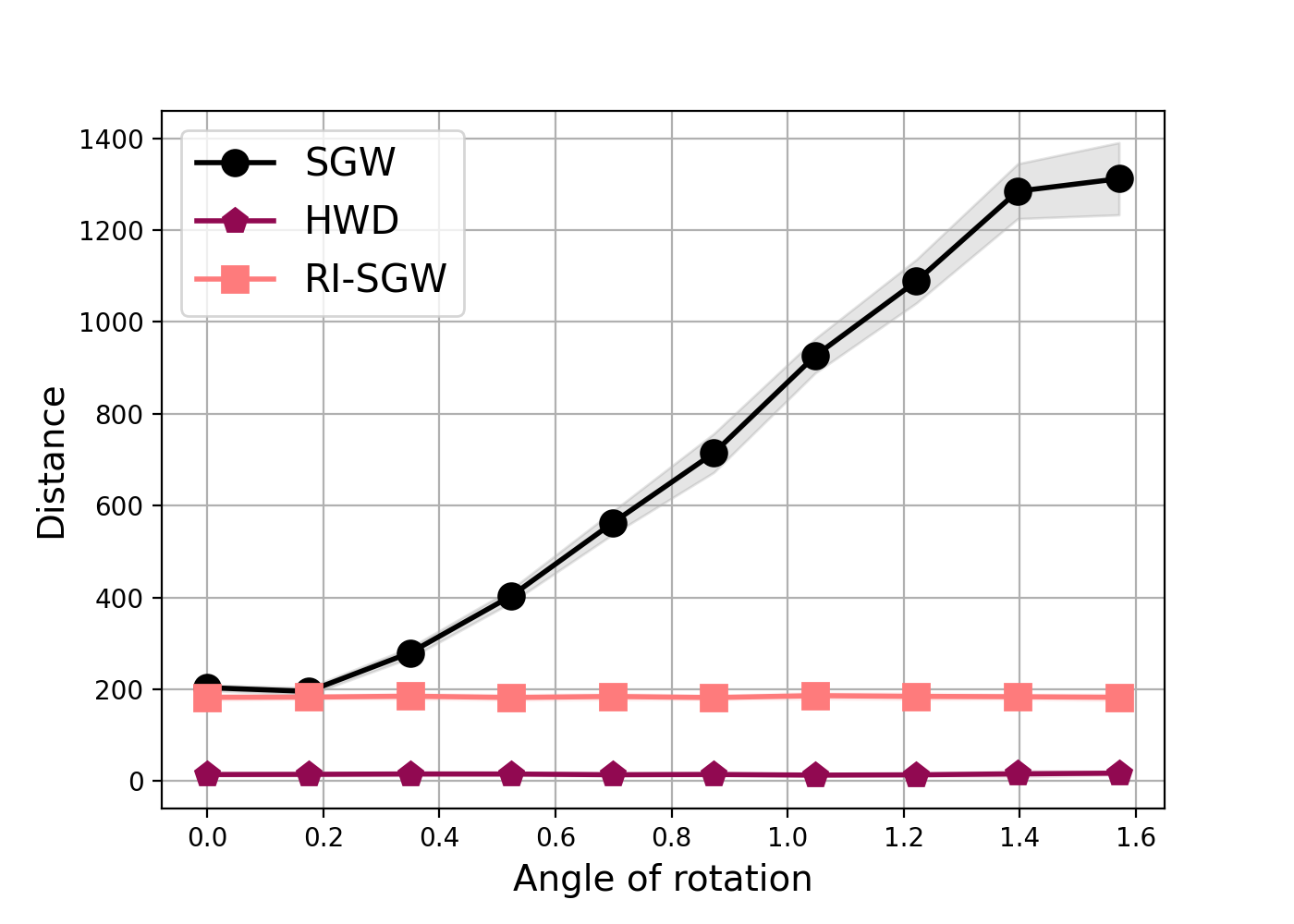}~\hfill~
\caption{Examples of distance computation between (left) two-translating Gaussian distributions. (right) two spirals. \label{fig:rota}}
\end{minipage}
%\begin{min}
\hspace{0.3 in}
\begin{minipage}[b]{0.4\linewidth}
	~\hfill	\includegraphics[width=0.4\textwidth]{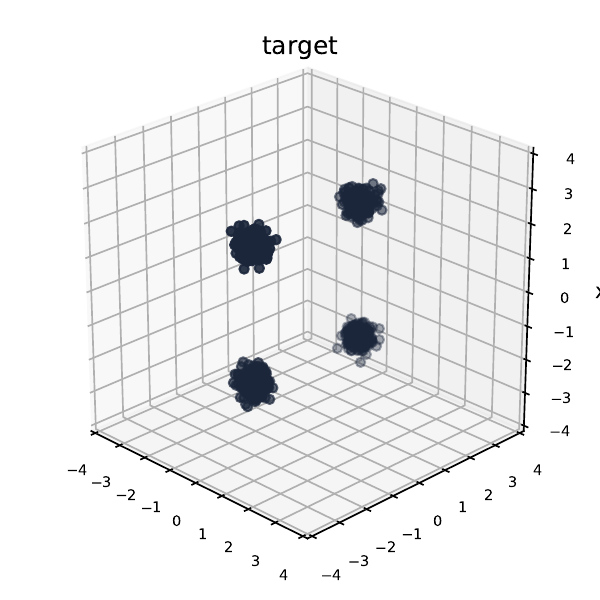}~\hfill~
	\includegraphics[width=0.4\textwidth]{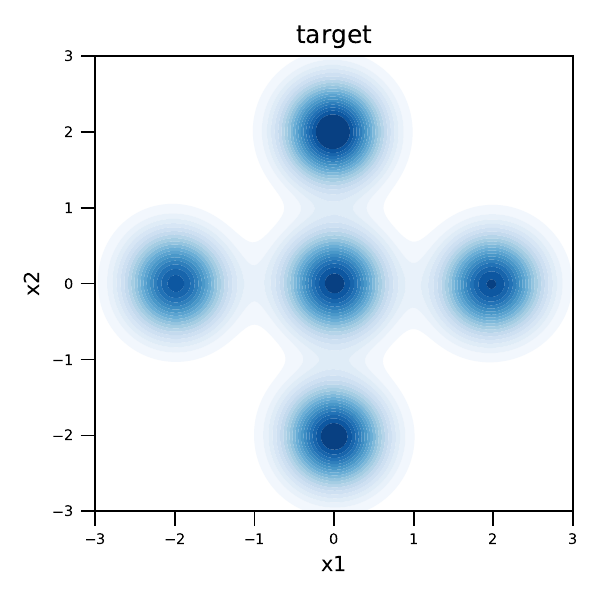}~\hfill~
	\caption{Examples of target distributions for our generative models (left) 3D $4$-mode. (right) 2D $5$-mode. \label{fig:gantarget}}
\end{minipage}

\end{figure}

 We have used two simple datasets for showing the behavior of HWD with respect to translation and rotation.
For translation, we consider two 2D Gaussian distributions one being fixed, the other with varying mean. For rotation, we use two 2D spirals from Scikit-Learn library, one being fixed and the other being rotated from $0$ to $\pi/2$. For these two cases, we have drawn $500$ samples, used $100$ random directions for SGW and RI-SGW. For our HWD, we have used only $10$ slices
% and $T=50, N=5$ iterations of the alternate optimization and $5$ iterations (resp. corresponding to variables $t$ and $i_k$ in Algorithm~\ref{algo:computeDSSE}).
and $T=50, N=5$ iterations for each of the alternate optimization.
The results we obtain are depicted in Figure~\ref{fig:rota}. From the first
panel, we remark that both SGW and RI-SGW are indeed insensitive to translation while HWD captures this translation, which is also verified by property $(vi)$ in Proposition~\ref{prop:dsse-distance}.
For the spiral problem,
as expected HWD and RI-SGW are indeed rotation-invariant while SGW is not. %Regarding computational running time, as one may have expected, SGW is about an order of magnitude more efficient than HWD which is itself
%an order of magnitude more efficient than RI-SGW.

%\AR{add results; 50000+100000 comment on complexity}

\paragraph{Generative models} 

\begin{figure}[htbp]%width=2.3mcm
\begin{center}
	%	~\hfill~
	\includegraphics[width=0.158\textwidth]{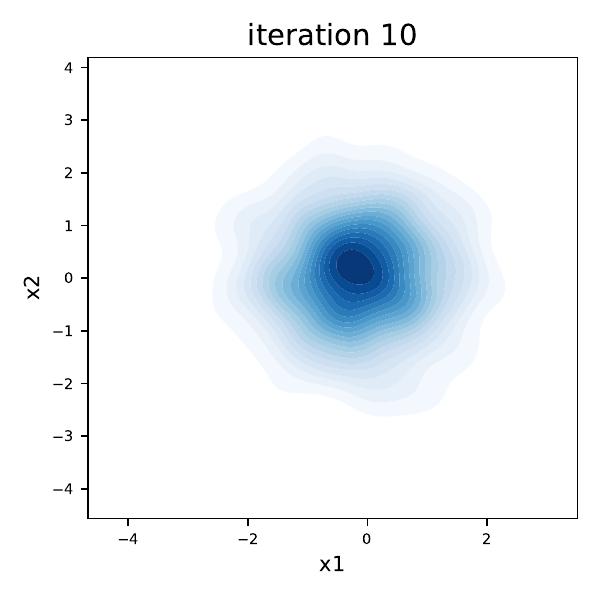}
	\includegraphics[width=0.158\textwidth]{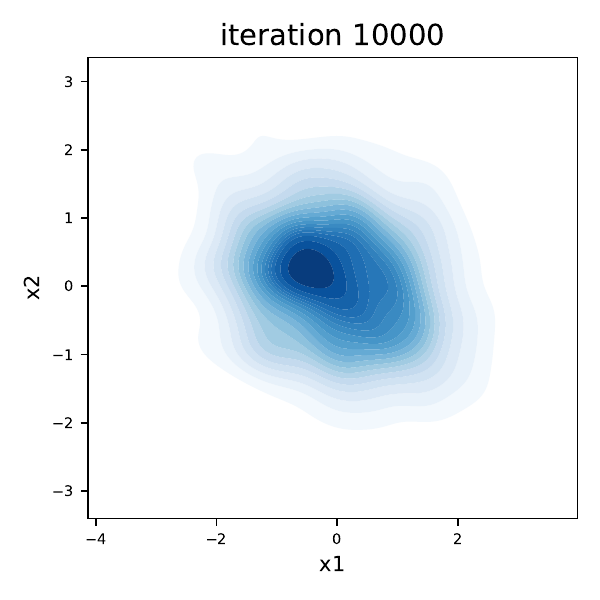}
	\includegraphics[width=0.158\textwidth]{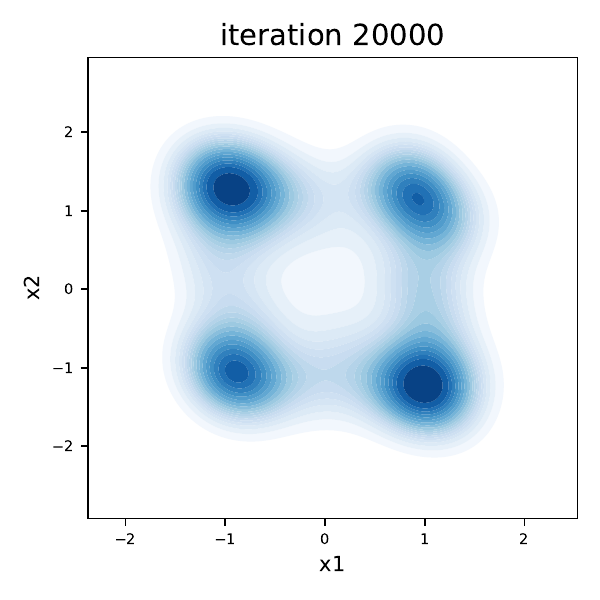}
	\includegraphics[width=0.158\textwidth]{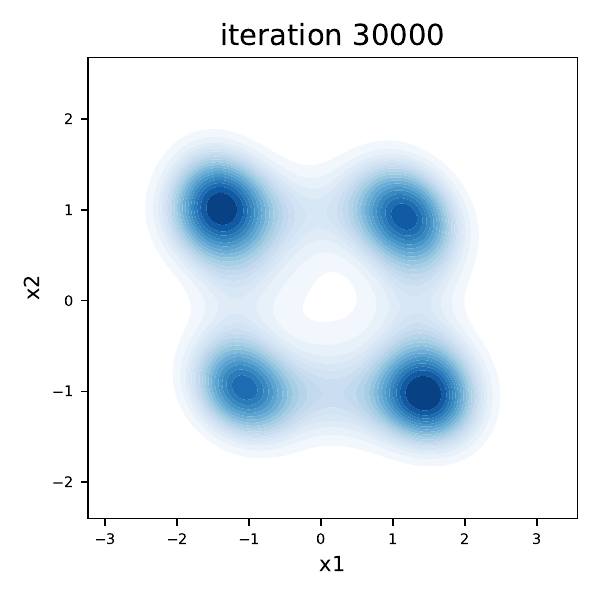}
	\includegraphics[width=0.158\textwidth]{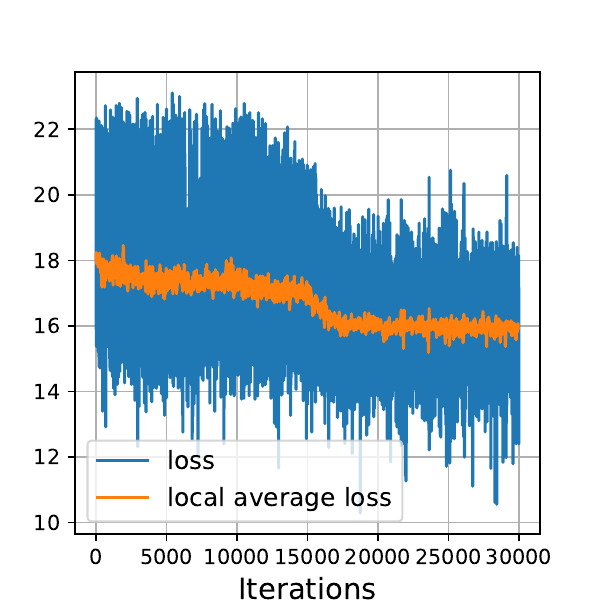} \\
	%~\hfill~
	\includegraphics[width=0.158\textwidth]{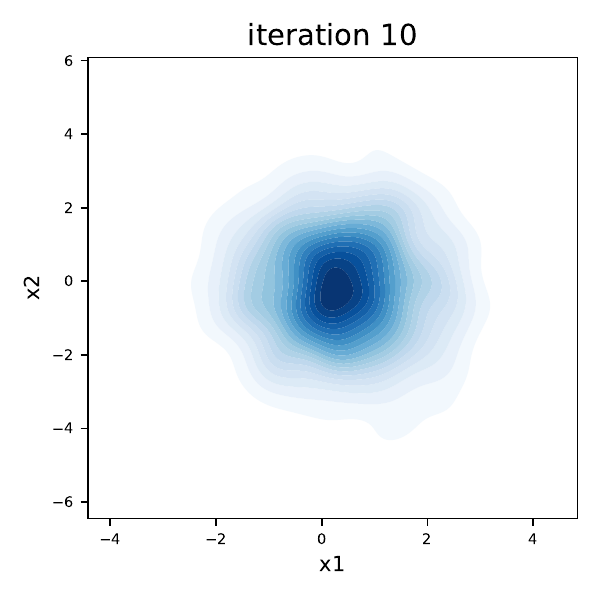}
	\includegraphics[width=0.158\textwidth]{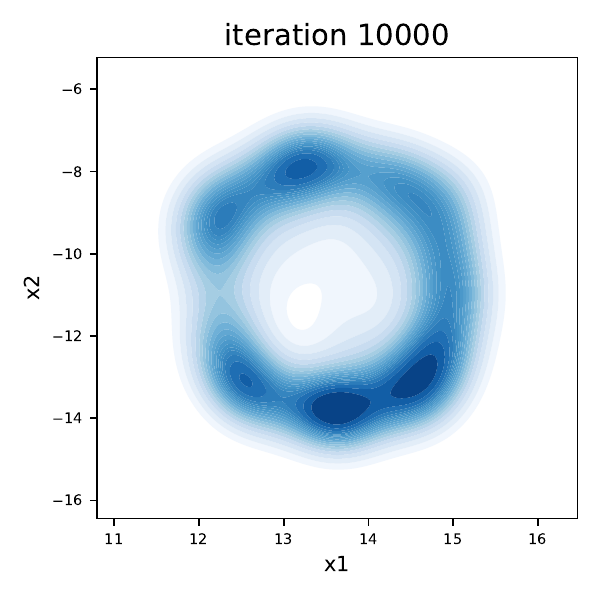}
	\includegraphics[width=0.158\textwidth]{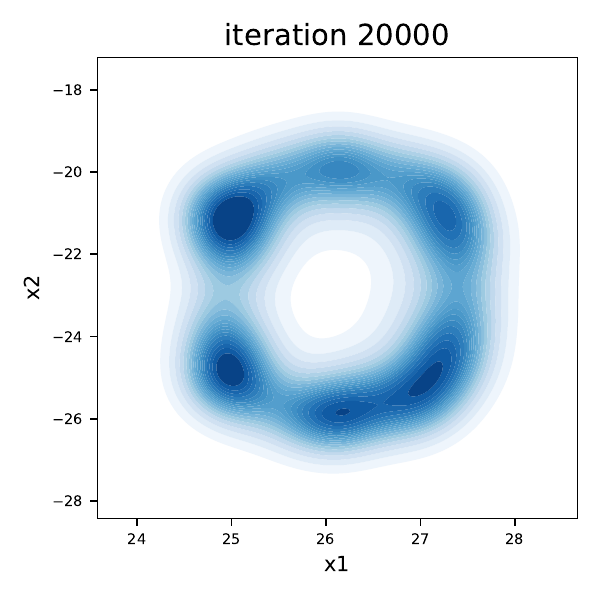}
	\includegraphics[width=0.158\textwidth]{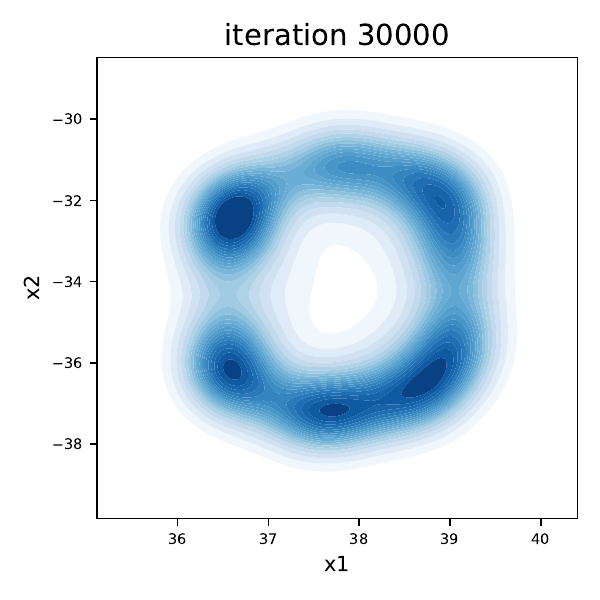}
	\includegraphics[width=0.158\textwidth]{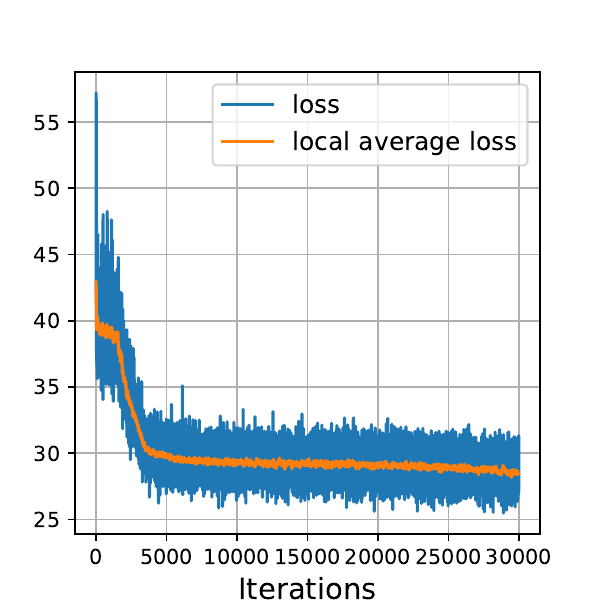}\\
	%~\hfill~
	\includegraphics[width=0.158\textwidth]{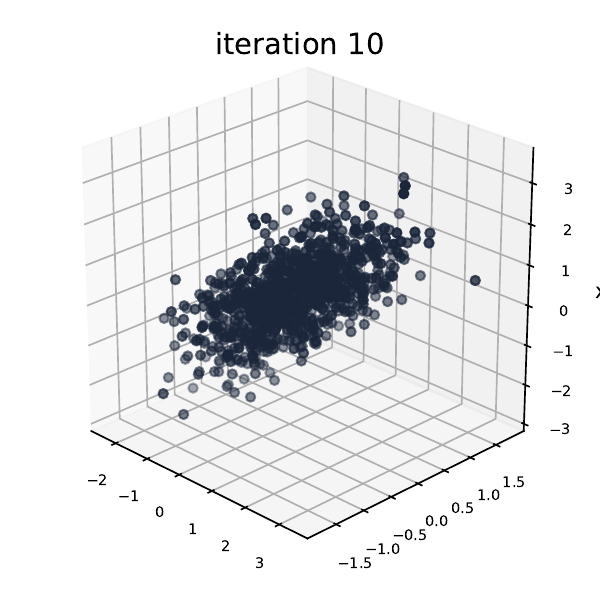}
	\includegraphics[width=0.158\textwidth]{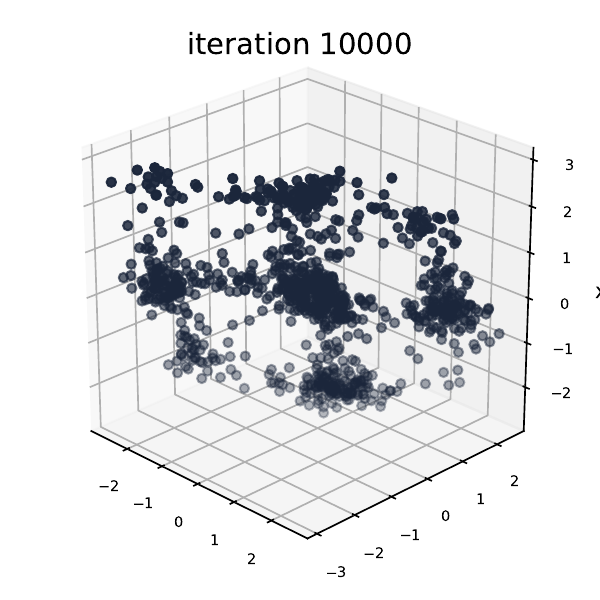}
	\includegraphics[width=0.158\textwidth]{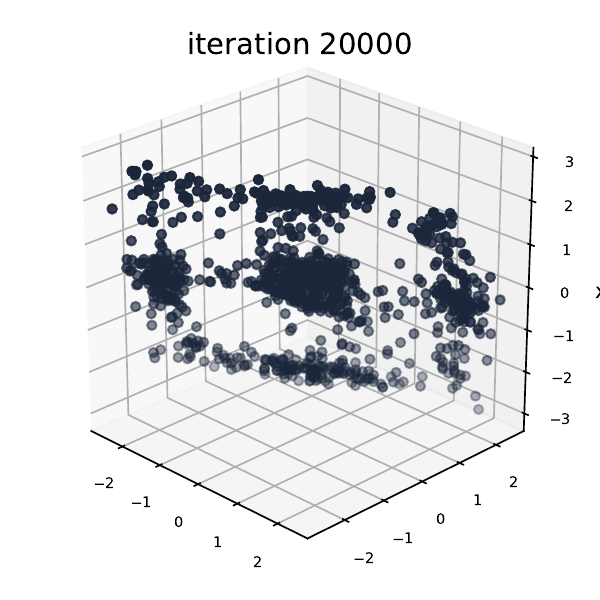}
	\includegraphics[width=0.158\textwidth]{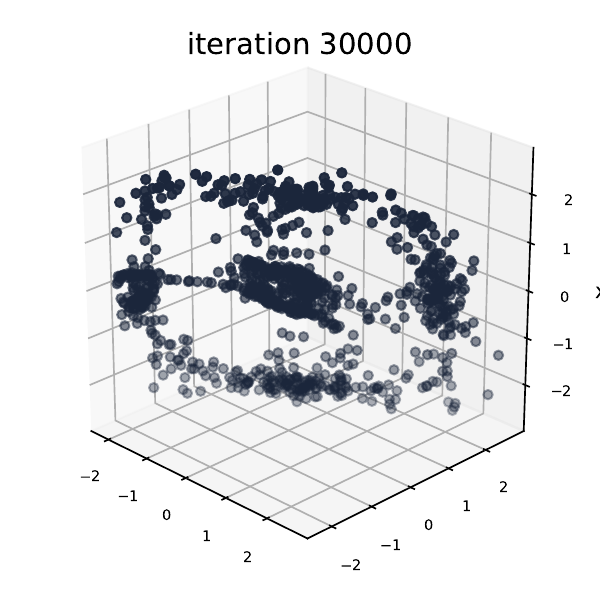}
	\includegraphics[width=0.158\textwidth]{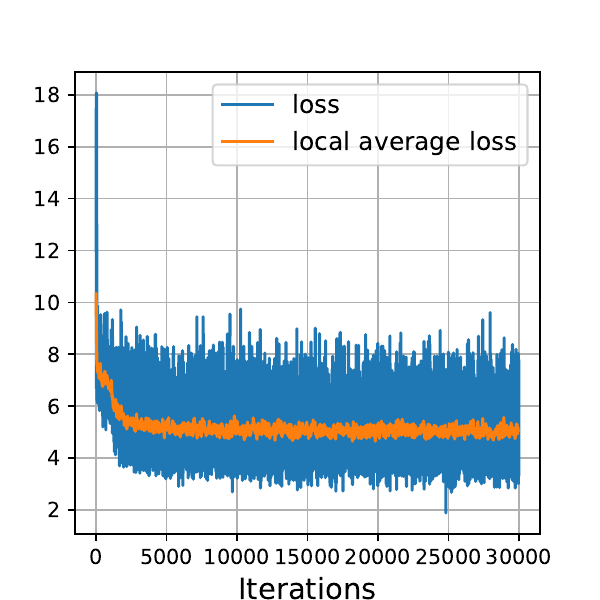} \\
	%~\hfill~ \\
	%~\hfill~
	\includegraphics[width=0.158\textwidth]{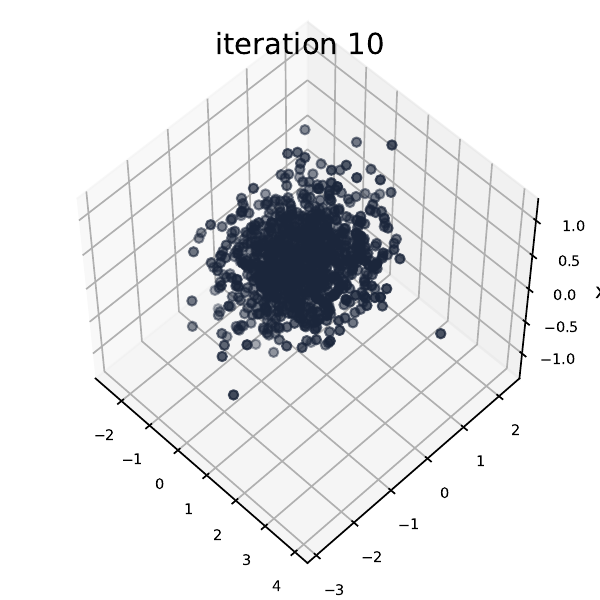}
	\includegraphics[width=0.158\textwidth]{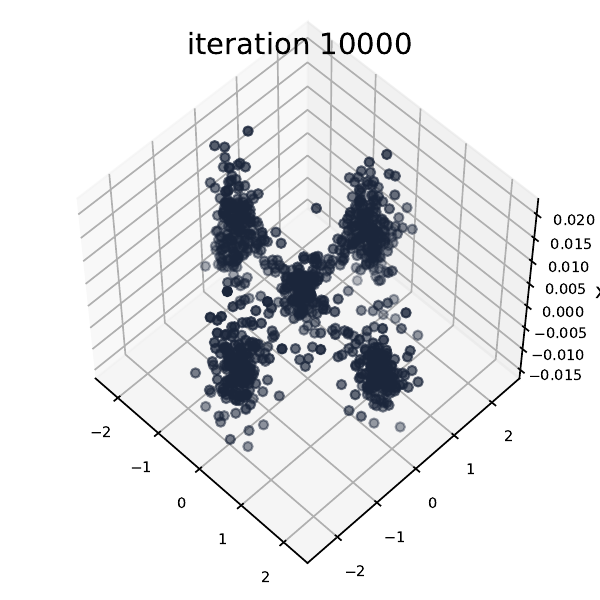}
	\includegraphics[width=0.158\textwidth]{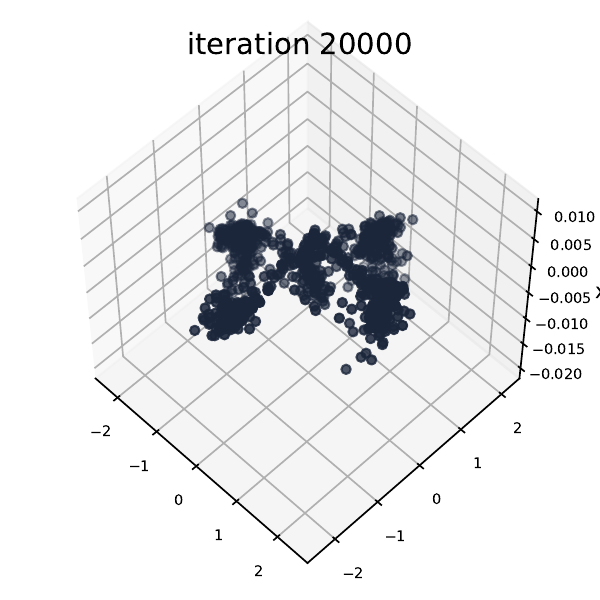}
	\includegraphics[width=0.158\textwidth]{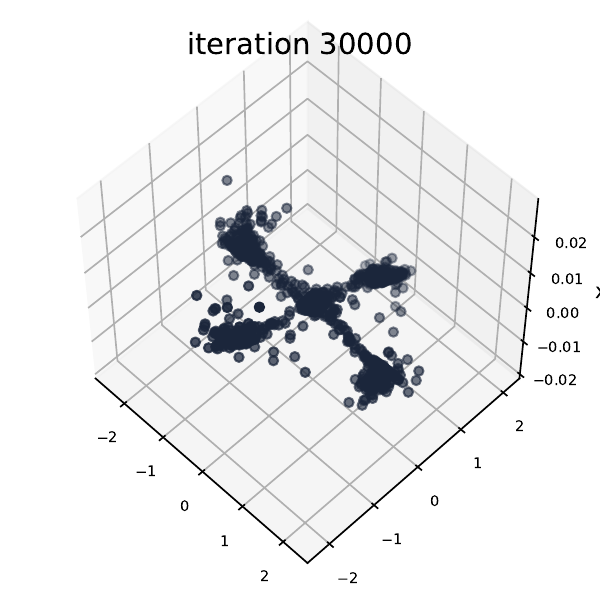}
	\includegraphics[width=0.158\textwidth]{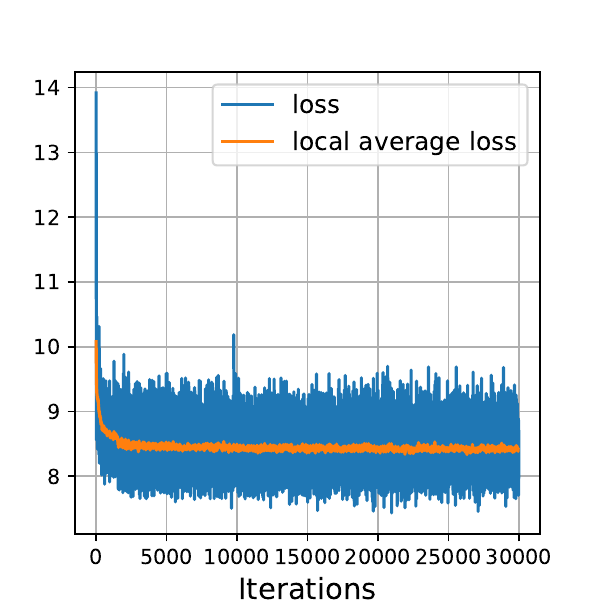}
	%~\hfill~
	 \\
	\caption{Examples of generated distributions across iterations $(10,10000, 20000,$ and $30000)$ for two targets.From top to bottom (first-row) HWD for the $4$-mode. (second-row) SGW for the $4$-mode (third-row) HWD for 3D $5$-mode. (fourth-row) SGW for $5$-mode.  For each row, the last panel shows the evolution of the loss over the $30000$ iterations.
	\label{fig:gan}}
\end{center}
\end{figure}

For checking whether our distribution discrepancy
behaves appropriately, we have used it as a loss function in a generative model.
Our task here is to build a model able to generate a distribution defined on
a space having a different dimensionality from the target distribution space.
As such, we have considered the same toy problems as in \cite{pmlr-v97-bunne19a}  
and investigated two situations: generating 2D distributions from 
3D data and the other way around. The 3D target distribution is a Gaussian mixture
model with four modes while the 2D ones are $5$-mode.
Our generative model is composed of a fully-connected neural network with ReLU activation functions. We have considered $3000$ samples in the target distributions
and batch size of $300.$ For both HWD and SGW, we have run the algorithm for $30000$
iterations with an Adam optimizer, stepsize of $0.001$ and default $\beta$ parameters. 
For the $4$-mode problem, the generator is a MLP with $2$ layers while for the $5$-mode, as the problem is more complex, it has $3$ layers. In each case, we have $256$ units on the first layer and then $128$. For the hyperparameters, we have set $\lambda_C=1$ and $\lambda_a=5$ or $\lambda_a=50$ depending on the problem. Note that for SGW, we have also added
a $\ell_2$-norm regularizer on the output of the generator in order to avoid
them to drift (see Figures~\ref{figure_8} and~\ref{figure_9} in Appendix~\ref{appendix_more_xp}), as the loss is translation-invariant. Examples of generated distributions are depicted in Figure \ref{fig:gan}. We remark that our HWD is able to produce visually correct distributions whereas SGW  struggles in generating the 4 modes
and its 3D $5$-mode is squeezed on its third dimension.

\paragraph{Scalability} 

\begin{figure}[ht]
	~\hfill
	\includegraphics[width=0.11\linewidth]{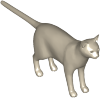} \hspace*{6pt}
	\includegraphics[width=0.11\linewidth]{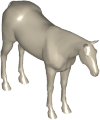}
	\includegraphics[width=0.11\linewidth]{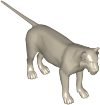} %\hspace*{12pt}
	~\hfill~
	\includegraphics[width=0.325\linewidth]{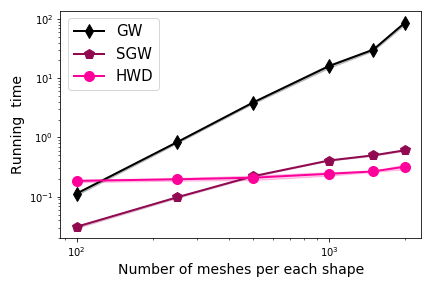}
	~\hfill
	\caption{Computation time with respect to $n$, the number of vertices on each shape. (Left-panel) Instances of 3D objects. (Right-panel) Running time.}
	\label{fig:running_time_shape_3d_3d}
\end{figure}

We consider the non-rigid shape world dataset \citep{bronstein2006efficient} which consists of $148$ three-dimensional shapes from $12$ classes. We draw randomly $n \in \left\{100,  250,  500, 1000, 1500, 2000 \right\}$ vertices $\{x_i \in \mathbb{R}^3\}_{i=1}^n$ on each shape and use them to measure the similarity between a pair of shapes $\{x_i \in \mathbb{R}^3\}_{i=1}^n$ and $\{y_j \in \mathbb{R}^3\}_{j=1}^n$. Figure \ref{fig:running_time_shape_3d_3d} reports the average time to compute on a single core such a similarity for $100$ pairs of shapes using respectively GW, SGW and HWD. As expected GW exhibits a slow behavior while the computational burden of HWD is on par with SGW.

%\paragraph{Classification under isometry transformations} 
\paragraph{Classification under various transformations} 
\begin{wrapfigure}{r}{0.3\textwidth}\vspace*{-4mm}
	\begin{center}
		\includegraphics[width=0.3\textwidth]{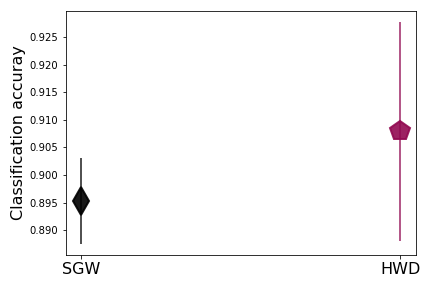}
	\end{center}
	\caption{Classification performance under transformations.}
	\label{fig:res_invarianceKNN}
\end{wrapfigure}

%This experiment, whose details are provided in the appendix, aims to evaluate the robustness of GW, SGW RI-SGW, and HWD to isometry in terms of classification accuracy. To that purpose we employ the Shape Retrieval Contest (SHREC'2010) correspondence dataset \cite{Bronstein_shrec2010}. It includes $3$ shape classes (see Figure~\ref{fig:robustnessshapes}) with $9$ different transformations and the null shape (no transformation). Each transformation is applied up to five  strength levels (weak to strong). We consider all strengths of the isometry transformation along with the null shape for each class leading to a number of $18$ samples. We perform a 1-NN classification. As the correspondence benchmark is of high resolution (10k-50K 3D meshes), we sample $n=1000$ vertices on each shape to compute the similarity matrix. Obtained performances over $10$ runs are depicted in Figure \ref{fig:res_invarianceKNN}. They highlight the ability of HWD to be robust to isometry perturbations. HWD achieves similar performances as GW but with far less computation time (see Figure \ref{fig:running_time_shape_3d_3d}).

This experiment, whose details are provided in Appendix~\ref{appendix-training_details}, aims to evaluate the robustness of  SGW and HWD (the computationally efficient methods) to different transformations in terms of classification accuracy. To that purpose we employ the Shape Retrieval Contest (SHREC'2010) correspondence dataset, see~\cite{Bronstein_shrec2010}. It includes high resolution (10K-50K) triangular meshes. The shapes are of $3$ classes (see Figure~\ref{fig:robustnessshapes} in Appendix~\ref{appendix_more_xp}) with $9$ different transformations and the null shape (no transformation). Each transformation is applied up to five  strength levels (weak to strong). Along with the null shape, we consider all strengths of the "isometry", "topology", "scale", "shotnoise" transformations %for each class 
leading to $63$ samples. We perform a 1-NN classification. 
%As the correspondence benchmark is of high resolution (10k-50K 3D meshes), we sample $n=1000$ vertices on each shape to compute the similarity matrix. 
Obtained performances over $10$ runs are depicted in Figure \ref{fig:res_invarianceKNN}. They highlight the ability of HWD to be robust to perturbations. HWD achieves slightly better mean classification accuracy than SGW with a competitive computation time (see Figure \ref{fig:running_time_shape_3d_3d}). Notice that GW and RISGW are unable to run under reasonable time-budget constraint.

% \newpage

\section{Conclusion} % (fold)
\label{sec:conclusion}
We introduce in this paper HWD a novel OT-based discrepancy between distributions lying in different spaces. It takes computational benefits from distributional slicing technique, which amounts to find an optimal number of random projections needed to capture the structure of data distributions. Another feature of this discrepancy consists in projecting the distributions in question through a learning of embeddings enjoying the same latent space. We showed a nice geometrical property verified by the proposed discrepancy, specifically a rotation-invariance. We illustrated through extensive experiments the applicability of this discrepancy on generative modeling and shape objects retrieval.
We argue that the implementation part faces the standard deep learning bottleneck of tuning the model's hyperparameters. %followed by a "min-max" problem in the objective loss of the discrepancy. 
A future extension line of this work is to deliver theoretical guarantees regarding the regularizing parameters, both of distributional and angle preserving properties.

% section conclusion (end)


\subsubsection*{Acknowledgments}
The works of Maxime Bérar, Gilles Gasso and Alain Rakotomamonjy have been supported by the OATMIL ANR-17-CE23-0012 Project of the French National Research Agency (ANR).

% \paragraph{Broader Impact}{

% Optimal transport tool has proven a significant usefulness for many tasks in machine learning or computer vision especially when dealing with distributions of real entities coming from different spaces. In that setting, this works combines distributional slicing and 1D-Wasserstein distance to propose an approximate discrepancy of distributions living in incomparable spaces. This discrepancy shares some main features of GW mainly the rotation invariance while exhibiting a favorable computation burden.  We expect the found results will prove useful for efficient computation of Sliced-GW and related applications. As is, we do not foresee any negative societal impacts of the proposed distributional sliced embedding (HWD) OT discrepancy. Rather, we show a computational gain compared with GW distance, and thus, we expect a better carbon-print footprint for applications based on HWD.}

\bibliography{biblio}

\begin{thebibliography}{40}
\providecommand{\natexlab}[1]{#1}
\providecommand{\url}[1]{\texttt{#1}}
\expandafter\ifx\csname urlstyle\endcsname\relax
  \providecommand{\doi}[1]{doi: #1}\else
  \providecommand{\doi}{doi: \begingroup \urlstyle{rm}\Url}\fi

\bibitem[Alaya et~al.(2020)Alaya, Bérar, Gasso, and
  Rakotomamonjy]{alaya2020theoretical}
M.~Z. Alaya, M.~Bérar, G.~Gasso, and A.~Rakotomamonjy.
\newblock Theoretical guarantees for bridging metric measure embedding and
  optimal transport, 2020.

\bibitem[Alvarez-Melis \& Jaakkola(2018)Alvarez-Melis and
  Jaakkola]{alvarezmelis2018gromov}
D.~Alvarez-Melis and T.~Jaakkola.
\newblock Gromov--{W}asserstein alignment of word embedding spaces.
\newblock In \emph{Proceedings of the 2018 Conference on Empirical Methods in
  Natural Language Processing}, pp.\  1881--1890. Association for Computational
  Linguistics, 2018.

\bibitem[Arjovsky et~al.(2017)Arjovsky, Chintala, and
  Bottou]{pmlr-v70-arjovsky17a}
M.~Arjovsky, S.~Chintala, and L.~Bottou.
\newblock {W}asserstein generative adversarial networks.
\newblock In Doina Precup and Yee~Whye Teh (eds.), \emph{Proceedings of the
  34th International Conference on Machine Learning}, volume~70 of
  \emph{Proceedings of Machine Learning Research}, pp.\  214--223,
  International Convention Centre, Sydney, Australia, 2017. PMLR.

\bibitem[Bonneel et~al.(2011)Bonneel, van~de Panne, Paris, and
  Heidrich]{bonnel2011}
N.~Bonneel, M.~van~de Panne, S.~Paris, and W.~Heidrich.
\newblock Displacement interpolation using lagrangian mass transport.
\newblock \emph{ACM Trans. Graph.}, 30\penalty0 (6):\penalty0 158:1--158:12,
  2011.

\bibitem[Bonneel et~al.(2015)Bonneel, Rabin, Peyr\'{e}, and
  Pfister]{bonneel-2015-sliced}
N.~Bonneel, J.~Rabin, G.~Peyr\'{e}, and H.~Pfister.
\newblock Sliced and {R}adon {W}asserstein barycenters of measures.
\newblock 51\penalty0 (1), 2015.

\bibitem[Bonnotte(2013)]{bonnotte:tel-00946781}
N.~Bonnotte.
\newblock \emph{{Unidimensional and Evolution Methods for Optimal
  Transportation}}.
\newblock Theses, {Universit{\'e} Paris Sud - Paris XI ; Scuola normale
  superiore (Pise, Italie)}, December 2013.

\bibitem[Bronstein et~al.(2010)Bronstein, Bronstein, Castellani, Falcidieno,
  Fusiello, Godil, Guibas, Kokkinos, Lian, Ovsjanikov, Patane, Spagnuolo, and
  Toldo]{Bronstein_shrec2010}
A.~Bronstein, M.~Bronstein, U.~Castellani, B.~Falcidieno, A.~Fusiello,
  A.~Godil, L.~Guibas, I.~Kokkinos, Z.~Lian, M.~Ovsjanikov, G.~Patane,
  M.~Spagnuolo, and R.~Toldo.
\newblock {SHREC 2010}: robust large-scale shape retrieval benchmark.
\newblock Eurographics Workshop on 3D Object Retrieval(2010), Norrköping, -1,
  2010-05-02 2010.

\bibitem[Bronstein et~al.(2006)Bronstein, Bronstein, and
  Kimmel]{bronstein2006efficient}
A.~M Bronstein, M.~M. Bronstein, and R.~Kimmel.
\newblock Efficient computation of isometry-invariant distances between
  surfaces.
\newblock \emph{SIAM Journal on Scientific Computing}, 28\penalty0
  (5):\penalty0 1812--1836, 2006.

\bibitem[Bunne et~al.(2019)Bunne, Alvarez-Melis, Krause, and
  Jegelka]{pmlr-v97-bunne19a}
C.~Bunne, D.~Alvarez-Melis, A.~Krause, and S.~Jegelka.
\newblock Learning generative models across incomparable spaces.
\newblock In Kamalika Chaudhuri and Ruslan Salakhutdinov (eds.),
  \emph{Proceedings of the 36th International Conference on Machine Learning},
  volume~97 of \emph{Proceedings of Machine Learning Research}, pp.\  851--861,
  Long Beach, California, USA, 09--15 Jun 2019. PMLR.

\bibitem[Chowdhury \& M{\'{e}}moli(2018)Chowdhury and
  M{\'{e}}moli]{chowdhury2018}
S.~Chowdhury and F.~M{\'{e}}moli.
\newblock The {G}romov--{W}asserstein distance between networks and stable
  network invariants.
\newblock \emph{CoRR}, abs/1808.04337, 2018.

\bibitem[Courty et~al.(2017)Courty, Flamary, Tuia, and
  Rakotomamonjy]{courty2017optimal}
N.~Courty, R.~Flamary, D.~Tuia, and A.~Rakotomamonjy.
\newblock Optimal transport for domain adaptation.
\newblock \emph{IEEE transactions on pattern analysis and machine
  intelligence}, 39\penalty0 (9):\penalty0 1853--1865, 2017.

\bibitem[Cuturi(2013)]{cuturinips13}
M.~Cuturi.
\newblock Sinkhorn distances: Lightspeed computation of optimal transport.
\newblock In C.~J.~C. Burges, L.~Bottou, M.~Welling, Z.~Ghahramani, and K.~Q.
  Weinberger (eds.), \emph{Advances in Neural Information Processing Systems
  26}, pp.\  2292--2300. Curran Associates, Inc., 2013.

\bibitem[Deshpande et~al.(2019)Deshpande, Hu, Sun, Pyrros, Siddiqui, Koyejo,
  Zhao, Forsyth, and Schwing]{max-SW}
I.~Deshpande, Y.-T. Hu, R.~Sun, A.~Pyrros, N.~Siddiqui, S.~Koyejo, Z.~Zhao,
  D.~Forsyth, and A.~G. Schwing.
\newblock Max-sliced {W}asserstein distance and its use for gans.
\newblock In \emph{2019 IEEE/CVF Conference on Computer Vision and Pattern
  Recognition (CVPR)}, pp.\  10640--10648, 2019.

\bibitem[Flamary et~al.(2021)Flamary, Courty, Gramfort, Alaya, Boisbunon,
  Chambon, Chapel, Corenflos, Fatras, Fournier, Gautheron, Gayraud, Janati,
  Rakotomamonjy, Redko, Rolet, Schutz, Seguy, Sutherland, Tavenard, Tong, and
  Vayer]{flamary2021pot}
R.~Flamary, N.~Courty, A.~Gramfort, M.~Z. Alaya, A.~Boisbunon, S.~Chambon,
  L.~Chapel, A.~Corenflos, K.~Fatras, N.~Fournier, L.~Gautheron, N.~T.H.
  Gayraud, H.~Janati, A.~Rakotomamonjy, I.~Redko, A.~Rolet, A.~Schutz,
  V.~Seguy, D.~J. Sutherland, R.~Tavenard, A.~Tong, and T.~Vayer.
\newblock Pot: Python optimal transport.
\newblock \emph{Journal of Machine Learning Research}, 22\penalty0
  (78):\penalty0 1--8, 2021.

\bibitem[Frogner et~al.(2015)Frogner, Zhang, Mobahi, Araya, and
  Poggio]{frogner2015nips}
C.~Frogner, C.~Zhang, H.~Mobahi, M.~Araya, and T.~A. Poggio.
\newblock Learning with a {W}asserstein loss.
\newblock In C.~Cortes, N.~D. Lawrence, D.~D. Lee, M.~Sugiyama, and R.~Garnett
  (eds.), \emph{Advances in Neural Information Processing Systems 28}, pp.\
  2053--2061. Curran Associates, Inc., 2015.

\bibitem[Gautschi(1959)]{gautchi-gamma}
W.~Gautschi.
\newblock Some elementary inequalities relating to the gamma and incomplete
  gamma function.
\newblock \emph{Journal of Mathematics and Physics}, 38\penalty0
  (1-4):\penalty0 77--81, 1959.

\bibitem[Kantorovich(1942)]{kantorovich1942}
L.~Kantorovich.
\newblock On the transfer of masses (in russian).
\newblock \emph{Doklady Akademii Nauk}, 2:\penalty0 227--229, 1942.

\bibitem[Kolouri et~al.(2017)Kolouri, Park, Thorpe, Slepcev, and
  Rohde]{klouri17}
S.~Kolouri, S.~R. Park, M.~Thorpe, D.~Slepcev, and G.~K. Rohde.
\newblock Optimal mass transport: Signal processing and machine-learning
  applications.
\newblock \emph{IEEE Signal Processing Magazine}, 34\penalty0 (4):\penalty0
  43--59, July 2017.

\bibitem[Kolouri et~al.(2019)Kolouri, Pope, Martin, and
  Rohde]{kolouri2018sliced}
S.~Kolouri, P.~E. Pope, C.~E. Martin, and G.~K. Rohde.
\newblock Sliced {W}asserstein auto-encoders.
\newblock In \emph{International Conference on Learning Representations}, 2019.

\bibitem[Kusner et~al.(2015)Kusner, Sun, Kolkin, and Weinberger]{kusnerb2015}
M.~Kusner, Y.~Sun, N.~Kolkin, and K.~Weinberger.
\newblock From word embeddings to document distances.
\newblock In Francis Bach and David Blei (eds.), \emph{Proceedings of the 32nd
  International Conference on Machine Learning}, volume~37 of \emph{Proceedings
  of Machine Learning Research}, pp.\  957--966, Lille, France, 07--09 Jul
  2015. PMLR.

\bibitem[Lee \& Sidford(2014)Lee and Sidford]{leeSidford2013PathFI}
Y.~T. Lee and A.~Sidford.
\newblock Path finding methods for linear programming: Solving linear programs
  in \~{O}(vrank) iterations and faster algorithms for maximum flow.
\newblock In \emph{Proceedings of the 2014 IEEE 55th Annual Symposium on
  Foundations of Computer Science}, FOCS '14, pp.\  424--433, Washington, DC,
  USA, 2014. IEEE Computer Society.

\bibitem[Lerner(2014)]{lerner2014course}
N.~Lerner.
\newblock \emph{A Course on Integration Theory}.
\newblock Springer Basel, 2014.

\bibitem[Lin et~al.(2021)Lin, Zheng, Chen, Cuturi, and
  Jordan]{pmlr-v130-lin21a}
T.~Lin, Z.~Zheng, E.~Chen, M.~Cuturi, and M.~Jordan.
\newblock On projection robust optimal transport: Sample complexity and model
  misspecification.
\newblock In Arindam Banerjee and Kenji Fukumizu (eds.), \emph{Proceedings of
  The 24th International Conference on Artificial Intelligence and Statistics},
  volume 130 of \emph{Proceedings of Machine Learning Research}, pp.\
  262--270. PMLR, 13--15 Apr 2021.

\bibitem[M{\'e}moli(2011)]{memoli2011GW}
F.~M{\'e}moli.
\newblock Gromov--{W}asserstein distances and the metric approach to object
  matching.
\newblock \emph{Foundations of Computational Mathematics}, 11\penalty0
  (4):\penalty0 417--487, 2011.

\bibitem[Monge(1781)]{monge1781}
G.~Monge.
\newblock Mémoire sur la théotie des déblais et des remblais.
\newblock \emph{Histoire de l'Académie Royale des Sciences}, pp.\  666--704,
  1781.

\bibitem[Nguyen et~al.(2020)Nguyen, Ho, Pham, and
  Bui]{nguyen2020distributional}
K.~Nguyen, N.~Ho, T.~Pham, and H.~Bui.
\newblock Distributional sliced-{W}asserstein and applications to generative
  modeling, 2020.

\bibitem[Paty \& Cuturi(2019)Paty and Cuturi]{patycuturi2019}
F.-P. Paty and M.~Cuturi.
\newblock Subspace robust {W}asserstein distances.
\newblock In Kamalika Chaudhuri and Ruslan Salakhutdinov (eds.),
  \emph{Proceedings of the 36th International Conference on Machine Learning},
  volume~97 of \emph{Proceedings of Machine Learning Research}, pp.\
  5072--5081, Long Beach, California, USA, 2019. PMLR.

\bibitem[Peyr{\'e} et~al.(2016)Peyr{\'e}, Cuturi, and
  Solomon]{peyre2016Gromov-W}
G.~Peyr{\'e}, M.~Cuturi, and J.~Solomon.
\newblock Gromov--{W}asserstein averaging of kernel and distance matrices.
\newblock In \emph{Proceedings of the 33rd International Conference on
  International Conference on Machine Learning - Volume 48}, ICML'16, pp.\
  2664--2672. JMLR.org, 2016.

\bibitem[Peyré \& Cuturi(2019)Peyré and Cuturi]{peyre2019COTnowpublisher}
G.~Peyré and M.~Cuturi.
\newblock Computational optimal transport.
\newblock \emph{Foundations and Trends® in Machine Learning}, 11\penalty0
  (5-6):\penalty0 355--607, 2019.

\bibitem[Rabin et~al.(2012)Rabin, Peyr{\'e}, Delon, and
  Bernot]{rabin-sliced-2011}
J.~Rabin, G.~Peyr{\'e}, J.~Delon, and M.~Bernot.
\newblock {W}asserstein barycenter and its application to texture mixing.
\newblock In A.~M. Bruckstein, B.~M. ter Haar~Romeny, A.~M. Bronstein, and
  M.~M. Bronstein (eds.), \emph{Scale Space and Variational Methods in Computer
  Vision}, pp.\  435--446, Berlin, Heidelberg, 2012. Springer Berlin
  Heidelberg.

\bibitem[Rachev \& R{\"u}schendorf(1998)Rachev and
  R{\"u}schendorf]{rachev1998mass}
S.T. Rachev and L.~R{\"u}schendorf.
\newblock \emph{Mass Transportation Problems: Volume I: Theory}.
\newblock Mass Transportation Problems. Springer, 1998.

\bibitem[Rowland et~al.(2019)Rowland, Hron, Tang, Choromanski, Sarlos, and
  Weller]{pmlr-v89-rowland19a}
M.~Rowland, J.~Hron, Y.~Tang, K.~Choromanski, T.~Sarlos, and A.~Weller.
\newblock Orthogonal estimation of {W}asserstein distances.
\newblock In Kamalika Chaudhuri and Masashi Sugiyama (eds.), \emph{Proceedings
  of the Twenty-Second International Conference on Artificial Intelligence and
  Statistics}, volume~89 of \emph{Proceedings of Machine Learning Research},
  pp.\  186--195. PMLR, 16--18 Apr 2019.

\bibitem[Salimans et~al.(2018)Salimans, Zhang, A.Radford, and
  Metaxas]{salimans2018improving}
T.~Salimans, H.~Zhang, A.Radford, and D.~Metaxas.
\newblock Improving {GAN}s using optimal transport.
\newblock In \emph{International Conference on Learning Representations}, 2018.

\bibitem[Solomon et~al.(2015)Solomon, de~Goes, Peyr{\'e}, Cuturi, Butscher,
  Nguyen, Du, and Guibas]{solomon2015}
J.~Solomon, F.~de~Goes, G.~Peyr{\'e}, M.~Cuturi, A.~Butscher, A.~Nguyen, T.~Du,
  and L.~Guibas.
\newblock Convolutional {W}asserstein distances: Efficient optimal
  transportation on geometric domains.
\newblock \emph{ACM Trans. Graph.}, 34\penalty0 (4):\penalty0 66:1--66:11,
  2015.

\bibitem[Sturm(2006)]{sturm2006}
K.~T. Sturm.
\newblock On the geometry of metric measure spaces. ii.
\newblock \emph{Acta Math.}, 196\penalty0 (1):\penalty0 133--177, 2006.

\bibitem[Vayer et~al.(2018)Vayer, Chapel, Flamary, Tavenard, and
  Courty]{vayer2018fsw}
T.~Vayer, L.~Chapel, R.~Flamary, R.~Tavenard, and N.~Courty.
\newblock Fused {G}romov--{W}asserstein distance for structured objects:
  theoretical foundations and mathematical properties.
\newblock \emph{CoRR}, abs/1811.02834, 2018.

\bibitem[Vayer et~al.(2019)Vayer, Flamary, Courty, Tavenard, and Chapel]{sgw}
T.~Vayer, R.~Flamary, N.~Courty, R.~Tavenard, and L.~Chapel.
\newblock Sliced {G}romov--{W}asserstein.
\newblock In H.~Wallach, H.~Larochelle, A.~Beygelzimer, F.~d~Alch\'{e}-Buc,
  E.~Fox, and R.~Garnett (eds.), \emph{Advances in Neural Information
  Processing Systems 32}, pp.\  14726--14736. Curran Associates, Inc., 2019.

\bibitem[Villani(2009)]{villani09optimal}
C.~Villani.
\newblock \emph{Optimal Transport: Old and New}, volume 338 of
  \emph{Grundlehren der mathematischen Wissenschaften}.
\newblock Springer Berlin Heidelberg, 2009.

\bibitem[Wu et~al.(2019)Wu, Huang, Acharya, Li, Thoma, Paudel, and
  Gool]{wu2019sliced}
J.~Wu, Z.~Huang, D.~Acharya, W.~Li, J.~Thoma, D.~P. Paudel, and L.~V. Gool.
\newblock Sliced {W}asserstein generative models, 2019.

\bibitem[Xu et~al.(2019)Xu, Luo, and Carin]{xu2019neurips}
H.~Xu, D.~Luo, and L.~Carin.
\newblock Scalable {G}romov--{W}asserstein learning for graph partitioning and
  matching.
\newblock In \emph{Advances in Neural Information Processing Systems 32: Annual
  Conference on Neural Information Processing Systems 2019, NeurIPS 2019, 8-14
  December 2019, Vancouver, BC, Canada}, pp.\  3046--3056, 2019.

\end{thebibliography}
\bibliographystyle{iclr2022_conference}

\newpage
\appendix
%!TEX root = main.tex

\section{Proofs} % (fold)
\label{sec:proofs}

\subsection{Proof of Proposition~\ref{prop:angle-preserving}} % (fold)
\label{sub:proof_of_prop:angle-preserving}

We use the following result:
\begin{lemma}
\label{lem:lemma-gamma}
[Theorem 3 in~\citep{nguyen2020distributional}]
For uniform measure $\sigma^{d-1}$ on the unit sphere $\mbS^{d-1}$, we have
\begin{equation*}
\int_{\mbS^{d-1}\times \mbS^{d-1}} |\inr{\theta, \theta'}| \diff \sigma^{d-1}(\theta) \diff \sigma^{d-1}(\theta') = \frac{\Gamma(d/2)}{\sqrt{\pi} \Gamma((d+1)/2)}.
\end{equation*}
\end{lemma}
Hence, 
\begin{align*}
\int_{\mbS^{d-1}\times \mbS^{d-1}} |\inr{\phi(\theta), \phi(\theta')}| \diff \sigma^{d-1}(\theta) \diff \sigma^{d-1}(\theta')  &\stackrel{\textrm{(Assumption~\ref{ass:ange-preserving})}}{\leq} L_\phi\int_{\mbS^{d-1}\times \mbS^{d-1}} |\inr{\theta, \theta'}| \diff \sigma^{d-1}(\theta) \diff \sigma^{d-1}(\theta')\\
&\stackrel{\textrm{(Lemma~\ref{lem:lemma-gamma})}}{\leq} \frac{L_\phi\Gamma(d/2)}{\sqrt{\pi} \Gamma((d+1)/2)}.
\end{align*}
Therefore, as long as the $(\phi, \psi)$-admissible constants $C_\phi \geq \frac{L_\phi\Gamma(d/2)}{\sqrt{\pi} \Gamma((d+1)/2)}$ and 
$C_\psi \geq \frac{L_\psi\Gamma(d/2)}{\sqrt{\pi} \Gamma((d+1)/2)}$, we have $\sigma^{d-1} \in \mathscr{M}_{C_{\phi}}\cap \mathscr{M}_{C_{\psi}}.$ Now using a Gautschi’s inequality~\citep{gautchi-gamma} for the Gamma function, it yields that $\frac{\Gamma(d/2)}{\sqrt{\pi} \Gamma((d+1)/2)} \geq \frac{1}{\sqrt{\pi (d+1)/2}} \geq 1/d.$ 
Let $\bar{\sigma} = \sum_{l=1}^d \frac 1d \delta_{\theta_l}$, where $\{\theta_1, \ldots, \theta_d\}$ form an orthonormal basis in $\R^d$. We then have 
\begin{align*}
\E_{\theta, \theta' \sim \bar{\sigma}}\big[|\inr{\phi(\theta), \phi(\theta')}|\big] = \sum_{1\leq k, l \leq d}\big(\frac 1d\big)^2 |\inr{\phi(\theta_k), \phi(\theta'_{l})}|% \\
&\stackrel{\textrm{(Assumption~\ref{ass:ange-preserving})}}{\leq} L_\phi \sum_{1\leq k, l \leq d}\big(\frac 1d\big)^2 |\inr{\theta_k, \theta'_{l}}|% \\
=\frac{L_\phi}{d}. 
\end{align*}
Therefore we get the lower bounds for 
the $(\phi, \psi)$-admissible constants $C_\phi$ and $C_\psi$ given in Proposition~\ref{prop:angle-preserving}, that guarantee $\sigma^{d-1}, \bar{\sigma} \in \mathscr{M}_{C_\phi}\cap\mathscr{M}_{C_{\psi}}.$

\subsection{Proof of Proposition~\ref{prop:dsse-distance}} % (fold)
\label{sub:proof_of_theorem_th:dsse-distance}

Let us first state the two following lemmas: 
Lemma~\ref{lem:integration_pushs} writes an integration result using push-forward measures; it relates integrals with respect to a measure $\eta$ and its push-forward under a measurable map $f: \cX \rightarrow \cY.$ Lemma~\ref{lem:image_admis_couplings} proves that the admissible set of couplings between the embedded measures are exactly the embedded of the admissible couplings between the original measures. 
\begin{lemma}
\label{lem:integration_pushs}
[See~\cite{lerner2014course} p. 61]
Let $f: S \rightarrow T$ be a measurable mapping, let $\eta$ be a measurable  measure on $S$, and let $g$ be  a measurable function on $T$. Then $\int_T g \diff f_{\#}\eta= \int_S (g\circ f) \diff\eta$.
\end{lemma}
\begin{lemma}
\label{lem:image_admis_couplings}
[Lemma  6 in~\cite{patycuturi2019}]
For all $\phi, \psi$ and $\mu \in \mathscr{P}(\cX), \nu \in \mathscr{P}(\cY)$, one has $\Pi(\phi{\#}\mu,\psi{\#}\nu) = \{(\phi \otimes \psi)\# \gamma \textrm{ s.t. } \gamma \in\Pi(\mu, \nu)\},$
% \begin{equation*}
% \Pi(\phi{\#}\mu,\psi){\#}\nu = \{(\phi \otimes \psi)\# \pi \textrm{ s.t. } \pi \in\Pi(\mu, \nu)\}
% \end{equation*}
where $\phi \otimes \psi: \cX \times \cY \rightarrow \cX \times \cY$ such that  $(\phi \otimes \psi(x,y) = (\phi(x), \psi(y))$ for all $x,y \in \cX \times \cY.$
\end{lemma}

$\bullet\,(i)$ {\it $\mathcal{HWD}_r(\mu, \mu)$ is finite.} In one hand, we assume that $\mu \in \mathscr{P}_r(\cX)$ and $\nu \in \mathscr{P}_r(\cY)$, hence its $r$-th moments are finite, i.e., $M_r(\mu) = \big(\int_\cX \norm{x}^r\diff \mu(x)\big)^{1/r} < \infty$ and $M_r(\nu) = \big(\int_\cY \norm{y}^r\diff \nu(y)\big)^{1/r} < \infty$. In the other hand, the following holds for all parameter $\theta \in \mbS^{d-1}$ and a couple $(\phi, \psi)$-embeddings, 
\begin{align*}
\mathcal{W}_r^r(\mu_{\phi, \theta}, \nu_{\psi, \theta})	&= \inf_{\pi \in \Pi(P_{\phi(\theta)}\#\mu,\psi(\theta)\#\nu)} \int_{\R\times \R} |u -u'|^r \diff \pi(u,u')\\
&\stackrel{\textrm{(Lemma~\ref{lem:image_admis_couplings})}}{=} \inf_{\gamma \in \Pi(\mu,\nu)} \int_{\cX\times \cY} |\phi(\theta)^\top x - \psi(\theta)^\top y|^r \diff \gamma(x,y)\\
&\leq 2^{r-1}\inf_{\gamma \in \Pi(\mu,\nu)} \int_{\cX\times \cY} \big(|\phi(\theta)^\top x|^r + |\psi(\theta)^\top y|^r\big) \diff \gamma(x,y)\\
&= 2^{r-1}\inf_{\gamma \in \Pi(\mu,\nu)} \Big(\int_{\cX} |\phi(\theta)^\top x|^r \diff \mu(x)  + \int_{\cY}|\psi(\theta)^\top y|^r \diff \nu(y)\Big), 
\end{align*}
where we use the facts that $(s+t)^r \leq 2^{r-1}(s^r + t^r), \forall s, t \in \R_+$ and that any $\gamma$ transport plan has marginals $\mu$ on $\cX$ and $\nu$ on $\cY$. By Cauchy–Schwarz inequality, we get 
\begin{align*}
\mathcal{W}_r^r(\mu_{\phi, \theta}, \nu_{\psi, \theta})	&\leq 2^{r-1} \Big(\int_{\cX} \norm{\phi(\theta)}^r \norm{x}^r\diff \mu(x) + \int_{\cY} \norm{\psi(\theta)}^r \norm{y}^r\diff \nu(y)\Big) = 2^{r-1} \big(M_r^r(\mu) + M_r^r(\nu)\big). 
\end{align*}
Then, $\big(\E_{\theta \sim \Theta}\big[\mathcal{W}_r^r(\mu_{\phi, \theta}, \nu_{\psi, \theta})\big]\big)^{1/r} \leq 2^{\frac{r-1}{r}} \big(M_r^r(\mu) + M_r^r(\nu)\big)^{1/r} \leq 2^{\frac{r-1}{r}}\big(M_r(\mu) + M_r(\nu)\big).$ Finally, one has that $\mathcal{HWD}_r(\mu, \nu) \leq  2^{\frac{r-1}{r}}\big(M_r(\mu) + M_r(\nu)\big).$

$\bullet\,(ii)$ {\it Non-negativity and symmetry.} Together the non-negativity, symmetry of Wasserstein distance and the decoupling property of iterated infima (or principle of the iterated infima)
%principle of iterated supremum 
yield the non-negativity and symmetry of the distributional sliced sub-embedding distance.%\\

$\bullet\,(ii)$ $\mathcal{HWD}_r(\mu, \mu) = 0$.
Let $\phi$ and $\phi'$ two embeddings for projecting the same distribution $\mu$. Without loss of generality, we suppose that the corresponding $(\phi, \phi')$-admissible constants $C_\phi' \leq C_\phi$, hence $\mathscr{M}_{C_{\phi'}}\subseteq \mathscr{M}_{C_{\phi}}$. Using the fact that $\sup(A\cap B) \leq \sup A\wedge \sup B,$ (with $a\wedge b) = \min (a,b))$, se have, straightforwardly,
\begin{align*}
\mathcal{HWD}_r(\mu, \mu)
&= \inf_{\phi, \phi'}\sup_{\Theta \in \mathscr{M}_{C_\phi}\cap\mathscr{M}_{C_{\phi'}}}\Big(\E_{\theta \sim \Theta}\big[\mathcal{W}_r^r(\mu_{\phi, \theta}, \mu_{\phi', \theta})\big]\Big)^{\frac 1r}\\
&\leq \inf_{\phi, \phi'} \bigg(\sup_{\Theta \in \mathscr{M}_{C_\phi}}\Big(\E_{\theta \sim \Theta}\big[\mathcal{W}_r^r(\mu_{\phi, \theta}, \mu_{\phi', \theta})\big]\Big)^{\frac 1r}\wedge \sup_{\Theta \in \mathscr{M}_{C_{\phi'}}}\Big(\E_{\theta \sim \Theta}\big[\mathcal{W}_r^r(\mu_{\phi, \theta}, \mu_{\phi', \theta})\big]\Big)^{\frac 1r}\bigg)\\
&= \inf_{\phi}\inf_{\phi'}\bigg(\sup_{\Theta \in \mathscr{M}_{C_\phi}}\Big(\E_{\theta \sim \Theta}\big[\mathcal{W}_r^r(\mu_{\phi, \theta}, \mu_{\phi', \theta})\big]\Big)^{\frac 1r}\wedge \sup_{\Theta \in \mathscr{M}_{C_{\phi'}}}\Big(\E_{\theta \sim \Theta}\big[\mathcal{W}_r^r(\mu_{\phi, \theta}, \mu_{\phi', \theta})\big]\Big)^{\frac 1r}\bigg)\\
&\leq \inf_{\phi}\bigg(\sup_{\Theta \in \mathscr{M}_{C_\phi}}\Big(\E_{\theta \sim \Theta}\big[\mathcal{W}_r^r(\mu_{\phi, \theta}, \mu_{\phi, \theta})\big]\Big)^{\frac 1r} \wedge \sup_{\Theta \in \mathscr{M}_{C_{\phi'}}}\Big(\E_{\theta \sim \Theta}\big[\mathcal{W}_r^r(\mu_{\phi, \theta}, \mu_{\phi, \theta})\big]\Big)^{\frac 1r}\bigg)\\
&\leq \inf_{\phi}\sup_{\Theta \in \mathscr{M}_{C_\phi}}\Big(\E_{\theta \sim \Theta}\big[\mathcal{W}_r^r(\mu_{\phi, \theta}, \mu_{\phi, \theta})\big]\Big)^{\frac 1r}\\% (\text{since } \mathscr{M}^{\phi}_{C'} \subseteq \mathscr{M}_{C_\phi})\\
&=0.
\end{align*}

$\bullet\,(iii)$ One has ${\big(\frac{1}{d}\big)^{\frac 1r}\inf_{\phi, \psi} \max_{\theta \in \mbS^{d-1}} \mathcal{W}_r(\mu_{\phi, \theta}, \nu_{\psi, \theta}) \leq \mathcal{HWD}_r(\mu, \nu) \leq \inf_{\phi, \psi}\max_{\theta \in \mbS^{d-1}} \mathcal{W}_r(\mu_{\phi, \theta}, \nu_{\psi, \theta})}.$ 
Since $\mathscr{M}_{C_\phi} \cap \mathscr{M}_{C_\psi} \subset \mathscr{M}_{1}$ and $\mathcal{W}_r^r(\mu_{\phi, \theta}, \nu_{\psi, \theta}) \leq \max_{\theta \in \mbS^{d-1}}\mathcal{W}_r^r(\mu_{\phi, \theta}, \nu_{\psi, \theta})$ we find that 
\begin{align*}
\sup_{\Theta \in \mathscr{M}_{C_\phi}\cap\mathscr{M}_{C_\psi}}\Big(\E_{\theta \sim \Theta}\big[\mathcal{W}_r^r(\mu_{\phi, \theta}, \nu_{\psi, \theta})\big]\Big)^{\frac 1r}
 &\leq  \sup_{\Theta \in \mathscr{M}_{1}}\Big(\E_{\theta \sim \Theta}\big[\mathcal{W}_r^r(\mu_{\phi, \theta}, \nu_{\psi, \theta})\big]\Big)^{\frac 1r} \\
&\leq  \big(\max_{\theta \in \mbS^{d-1}}\mathcal{W}^r_r(\mu_{\phi, \theta}, \nu_{\psi, \theta})\big)^{1/r}\\
&\leq \max_{\theta \in \mbS^{d-1}}\mathcal{W}_r(\mu_{\phi, \theta}, \nu_{\psi, \theta}),
% \sup_{\Theta \in \mathscr{M}_{1}}\Big(\E_{\theta \sim \Theta}\big[\max_{\theta \in \mbS^{d-1}}\mathcal{W}_r^r(\mu_{\phi, \theta}, \nu_{\psi, \theta})\big]\Big)^{\frac 1r}\\
\end{align*}
which entails that $\mathcal{HWD}_r(\mu, \nu) \leq \inf_{\phi, \psi}\max_{\theta \in \mbS^{d-1}}\mathcal{W}_r(\mu_{\phi, \theta}, \nu_{\psi, \theta})$. Moreover, since the $(\phi, \psi)$-admissible constants $C_\phi$ and $C_\psi$ satisfy $C_\phi\geq \frac{U_\phi}{d}$ and $C_\psi\geq \frac{U_\psi}{d}$ hence $\bar{\sigma} = \sum_{l=1}^d \frac 1d \delta_{\theta_l} \in \mathscr{M}_{C_\phi}\cap\mathscr{M}_{C_\psi}$, where we set $\theta_1 = \argmax_{\theta \in \mbS^{d-1}}\mathcal{W}_r(\mu_{\phi, \theta}, \nu_{\psi, \theta})$. We then obtain 
\begin{align*}
\mathcal{HWD}_r(\mu, \nu)& \geq  \inf_{\phi, \psi} \Big(\E_{\theta \sim \bar{\sigma}}\big[\mathcal{W}_r^r(\mu_{\phi, \theta}, \nu_{\psi, \theta})\big]\Big)^{\frac 1r}\\
&= \inf_{\phi, \psi} \Big(\sum_{l=1}^d \frac{1}{d}\mathcal{W}_r^r(\mu_{\phi, \theta_l}, \nu_{\psi, \theta_l})\Big)^{\frac 1r}\\
&\geq \big(\frac 1d\big)^{1/r} \inf_{\phi, \psi}
\mathcal{W}_r(\mu_{\phi, \theta_1}, \nu_{\psi, \theta_1})\\
&= \big(\frac 1d\big)^{1/r} \inf_{\phi, \psi} \max_{\theta \in \mbS^{d-1}}\mathcal{W}_r(\mu_{\phi, \theta}, \nu_{\psi, \theta}).
\end{align*}

$\bullet\,(iv)$ {\it For $p=q$, HWD is upper bound by the distributional Wasserstein distance (DSW)~.} Let us first recall the  DSW distance: let $C>0$ and set $\mathscr{M}_{C} =\{\Theta \in \mathscr{P}(\mbS^{d-1}): \E_{\theta, \theta' \sim \Theta}[|\inr{\theta, \theta'}|] \leq C\}.$
\begin{align*}
\mathcal{DSW}_r(\mu, \nu) = \sup_{\Theta \in \mathscr{M}_{C}}\Big(\E_{\theta \sim \Theta}\big[\mathcal{W}_r^r(\mu_{\theta}, \nu_{ \theta})\big]\Big)^{\frac 1r}.
\end{align*}
We have that the case of a identity couple of embeddings, $\phi = Id, \psi=Id$, the probability measure set $\mathscr{M}_{C_\phi}, \mathscr{M}_{C_\psi}=\mathscr{M}_{C},$ then it is trivial that $\mathcal{HWD}_r(\mu, \nu) \leq \mathcal{DSW}_r(\mu, \nu).$ 
% \begin{align*}
% \mathcal{HWD}_r(\mu, \nu) \leq 
% \end{align*}

$\bullet\,(v)$ {\it Rotation invariance.} 
Note that $(R\#\mu)_{\phi, \theta} = P_{\phi(\theta)}\#(R\#\mu)= (P_{\phi(\theta)}\circ R)\#\mu,$
% \begin{align*}
% (R\#\mu)_\theta^\phi &= P_{\phi(\theta)}\#(R\#\mu)= (P_{\phi(\theta)}\circ R)\#\mu,
% \end{align*}
and for all $x \in \R^p,$ using the adjoint operator $R^*$, $(R^* = R^{-1}$), $(P_{\phi(\theta)}\circ R) (x) = \inr{\phi(\theta), R(x)} = \inr{R^*(\phi(\theta)), x} = P_{R^* \circ \phi(\theta)}(x).$
% \begin{align*}
% (P_{\phi(\theta)}\circ R) (x) = \inr{\phi(\theta), R(x)} = \inr{R^*(\phi(\theta)), x} = P_{R^* \circ \phi(\theta)}(x).
% \end{align*}
Then, $(R\#\mu)_{\phi, \theta} = (P_{R^* \circ \phi(\theta)})\#\mu$. Analogously, one has $(Q\#\nu)_{\psi, \theta} = (P_{Q^* \circ \psi(\theta)})\#\nu$. Moreover, 
\begin{align*}
\mathscr{M}_{C_\phi} &= \big\{\Theta \in \mathscr{P}(\mbS^{d-1}):\E_{\theta, \theta' \sim \Theta}[|\inr{\phi(\theta), \phi(\theta')}|]\big\}\\
&= \big\{\Theta \in \mathscr{P}(\mbS^{d-1}):\E_{\theta, \theta' \sim \Theta}[|\inr{(R^*\circ\phi)(\theta), (R^*\circ\phi)(\theta')}|] \big\}\\
&= \mathscr{M}_{C_{R^* \circ \phi}}.
\end{align*}
Then we have similarly $\mathscr{M}_{C_\psi} = \mathscr{M}_{C_{Q^* \circ \psi}}$. %, for all $R, Q$ orthogonal mapping. 
This implies
\begin{align*}
\mathcal{HWD}_r(R\#\mu, Q\#\nu)&= \inf_{\phi, \psi}\sup_{\Theta \in \mathscr{M}_{C_\phi}\cap\mathscr{M}_{C_\psi}}\Big(\E_{\theta \sim \Theta}\big[\mathcal{W}_r^r((R\#\mu)_{\phi, \theta}, (Q\#\nu)_{\psi, \theta})\big]\Big)^{\frac 1r}\\
&= \inf_{\phi, \psi}\sup_{\Theta \in \mathscr{M}_{C_\phi}\cap\mathscr{M}_{C_\psi}}\Big(\E_{\theta \sim \Theta}\big[\mathcal{W}_r^r((P_{R^*\circ \phi(\theta)})\#\mu, P_{Q^* \circ \psi(\theta)})\#\nu)\big]\Big)^{\frac 1r}\\
&= \inf_{\phi, \psi}\sup_{\Theta \in \mathscr{M}_{C_\phi}\cap\mathscr{M}_{C_\psi}}\Big(\E_{\theta \sim \Theta}\big[\mathcal{W}_r^r(\mu_{{R^* \circ \phi},\theta}, \mu_{{R^* \circ \phi},\theta}\big]\Big)^{\frac 1r}\\
&= \inf_{\phi, \psi}\sup_{\Theta \in \mathscr{M}_{C_{R^* \circ \phi}}\cap\mathscr{M}_{C_{Q^* \circ \psi}}}\Big(\E_{\theta \sim \Theta}\big[\mathcal{W}_r^r(\mu_{{R^* \circ \phi},\theta}, \nu_{{Q^* \circ \phi},\theta}\big]\Big)^{\frac 1r}\\
&= \inf_{\phi, \psi}\sup_{\Theta \in \mathscr{M}_{C_{R^* \circ \phi}}\cap\mathscr{M}_{C_{Q^* \circ \psi}}}\big(\E_{\theta \sim \Theta}\big[\mathcal{W}_r^r(\mu_{{R^* \circ \phi}, \theta}, \nu_{{Q^* \circ \psi},\theta}\big]\big)^{\frac 1r}\\
&= \inf_{\phi'= R^* \circ \phi, \psi'= Q^* \circ \psi}\sup_{\Theta \in \mathscr{M}_{C_{\phi'}}\cap \mathscr{M}_{C_{\psi'}}}\Big(\E_{\theta \sim \Theta}\big[\mathcal{W}_r^r(\mu_{{\phi'},\theta}, \nu_{{\psi'},\theta}\big]\Big)^{\frac 1r}\\
&=\mathcal{HWD}_r(\mu, \nu).
\end{align*}
$\bullet\,(vi)$ {\it Translation quasi-invariance.} 
We have 
\begin{align*}
\mathcal{HWD}_r(T_\alpha\#\mu, T_\beta\#\nu)&= \inf_{\phi, \psi}\sup_{\Theta \in \mathscr{M}_{C_\phi}\cap\mathscr{M}_{C_\psi}}\Big(\E_{\theta \sim \Theta}\big[\mathcal{W}_r^r((T_\alpha\#\mu)_{\phi,\theta}, (T_\beta\#\nu)_{\psi,\theta})\big]\Big)^{\frac 1r}.
\end{align*}
By Lemmas~\ref{lem:image_admis_couplings} and~\ref{lem:integration_pushs} , we have 
\begin{align*}
\mathcal{W}_r^r(&(T_\alpha\#\mu)_{\phi,\theta}, (T_\beta\#\nu)_{\psi,\theta}\\
 &= \inf_{\gamma \in \Pi((T_\alpha\#\mu)_{\phi,\theta}, (T_\beta\#\nu)_{\psi,\theta}))} \int_{\R^2} |u - v|^r\diff \gamma(u,v)\\
&= \inf_{\gamma \in \Pi((P_{\phi(\theta)} \circ T_\alpha)\#\mu, (P_{\psi(\theta)}\circ T_\beta)\#\nu)} \int_{\R^2} |u - v|^r\diff \gamma(u,v)\\
&= \inf_{\gamma \in \Pi(\mu, \nu)} \int_{\cX\times\cY} |P_{\phi(\theta)} \circ T_\alpha (x) - P_{\psi(\theta)}\circ T_\beta)(y)|^r\diff \gamma(x,y)\\
&= \inf_{\gamma \in \Pi(\mu, \nu)} \int_{\cX\times\cY} |(P_{\phi(\theta)}(x) - P_{\psi(\theta)}(y)) +(P_{\phi(\theta)}(\alpha) - P_{\psi(\theta)}(\beta))|^r\diff \gamma(x,y)\\
&\leq 2^{r-1}\Big(\inf_{\gamma \in \Pi(\mu, \nu)} \int_{\cX\times\cY} |(P_{\phi(\theta)}(x) - P_{\psi(\theta)}(y))|^r\diff \gamma(x,y) + |P_{\phi(\theta)}(\alpha) - P_{\psi(\theta)}(\beta))|^r\Big)\\
&\leq 2^{r-1}\Big(\inf_{\gamma \in \Pi(\mu, \nu)} \int_{\cX\times\cY} |(P_{\phi(\theta)}(x) - P_{\psi(\theta)}(y))|^r\diff \gamma(x,y) + (\norm{\alpha} + \norm{\beta})^r\Big).
\end{align*}
% where in the last two inequalities, we use the facts that $(s+t)^r \leq 2^{r-1}(s^r + t^r), \forall s, t \in \R_+$ and a Cauchy–Schwarz inequality.  
Thanks to Minkowski inequality, 
\begin{align*}
&\sup_{\Theta \in \mathscr{M}_{C_\phi}\cap\mathscr{M}_{C_\psi}}\Big(\E_{\theta \sim \Theta}\big[\mathcal{W}_r^r((T_\alpha\#\mu)_\theta^\phi, (T_\beta\#\nu)_\theta^\psi) \big]\Big)^{\frac 1r}\\
& \leq 2^{r-1}\sup_{\Theta \in \mathscr{M}_{C_\phi}\cap\mathscr{M}_{C_\psi}}\Big(\E_{\theta \sim \Theta}\Big[\inf_{\gamma \in \Pi(\mu, \nu)} \int_{\cX\times\cY} |(P_{\phi(\theta)}(x) - P_{\psi(\theta)}(y))|^r\diff \gamma(x,y)\Big]\Big)^{\frac 1r}\\
&\qquad  + 2^{r-1}(\norm{\alpha} + \norm{\beta})\sup_{\Theta \in \mathscr{M}_{C_\phi}\cap\mathscr{M}_{C_\psi}}\big(\Theta(\mbS^{d-1})\big)^{\frac 1r}\\
& \leq 2^{r-1}\sup_{\Theta \in \mathscr{M}_{C_\phi}\cap\mathscr{M}_{C_\psi}}\Big(\E_{\theta \sim \Theta}\big[\mathcal{W}_r^r(\mu_{\phi, \theta}, \nu_{\psi,\theta})\big]\Big)^{\frac 1r} + 2^{r-1}(\norm{\alpha} + \norm{\beta}). %\sup_{\Theta \in \mathscr{M}_{C_\phi}\cap\mathscr{M}_{C_\psi}}\big(\Theta(\mbS^{d-1})\big)^{\frac 1r}.
%\sup_{\Theta \in \mathscr{M}_{C_\phi}\cap\mathscr{M}_{C_\psi}}\Big(\E_{\theta \sim \Theta}\big[|P_{\phi(\theta)}(\alpha) - P_{\psi(\theta)}(\beta))|^r\big]\Big)^{\frac 1r}.
% \sup_{\Theta \in \mathscr{M}^{\phi, \psi}_{C,K}}\big(
% \sup_{\Theta \in \mathscr{M}^{\phi, \psi}_{C,K}}\big(\E_{\theta \sim \Theta}\big[\inf_{\gamma \in \Pi(\mu, \nu)} \int_{\R^p\times\R^q} |(P_{\phi(\theta)}(x) - P_{\psi(\theta)}(y))|^r\diff \gamma(x,y)\big]\big)^{\frac 1r}\\
% &\qquad  + \sup_{\Theta \in \mathscr{M}^{\phi, \psi}_{C,K}}\big(\E_{\theta \sim \Theta}\big[|P_{\phi(\theta)}(\alpha) - P_{\psi(\theta)}(\beta))|\big]\big)\\
\end{align*}
Therefore, we get $\mathcal{HWD}_r(T_\alpha\#\mu, T_\beta\#\nu)\leq 2^{r-1}\mathcal{HWD}_r(\mu, \nu) + 2^{r-1}(\norm{\alpha} + \norm{\beta}).$

\section{Implementation} \label{sec:implementation}

This section graphically describes the learning procedure in Algorithm \ref{algo:computeDSSE}. It also provides the training details not exposed in the main body of the paper. %of the report results

\subsection{Learning scheme}

\begin{figure}[htbp]
\begin{center}
\includegraphics[width=11cm]{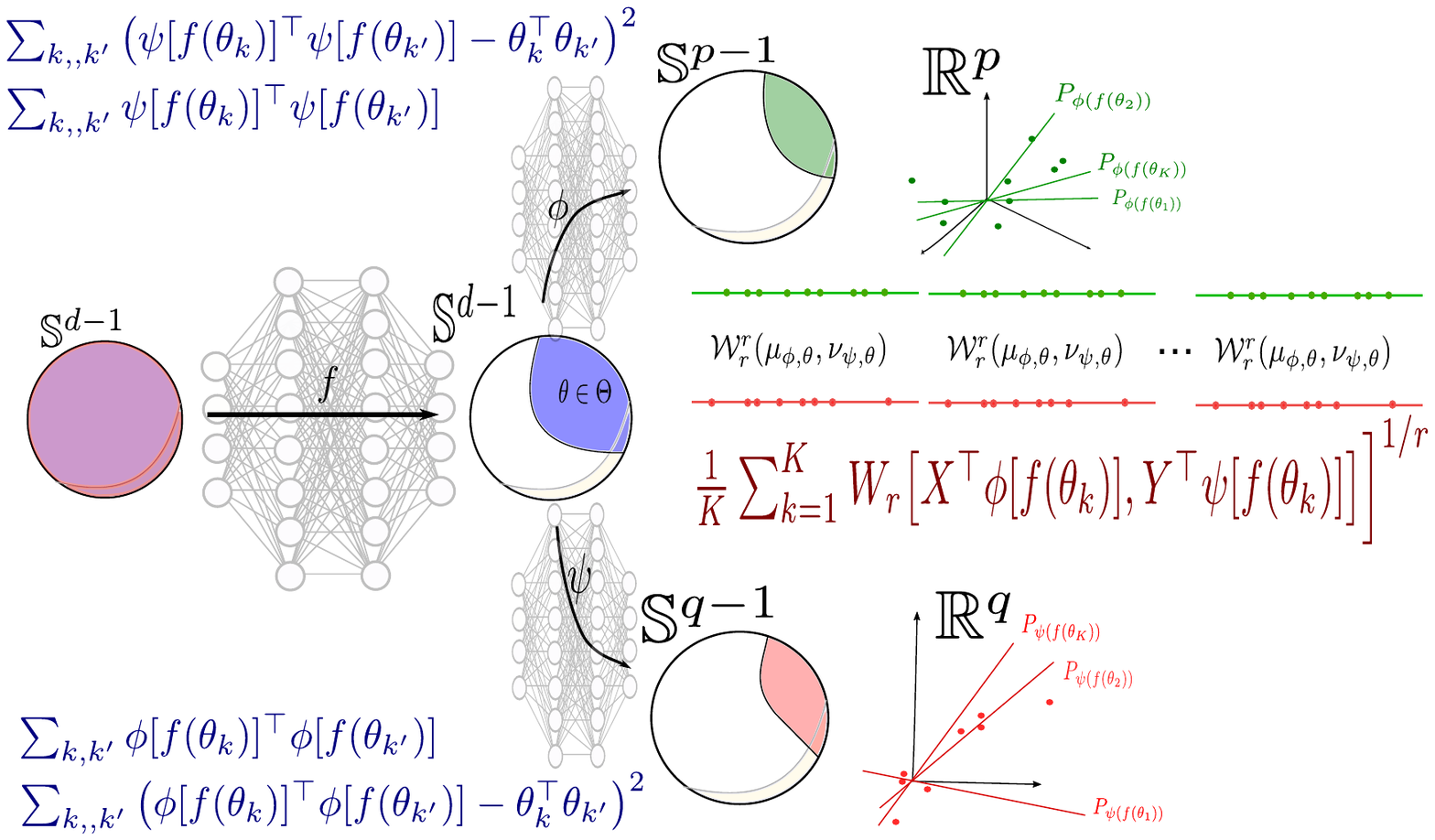}
	\end{center}
	\caption{The implemented approach. Both the distributional and mappings parts are achieved by deep neural networks. A number $K$ of projections is used to compute 1D-Wasserstein distances.}
	\label{fig:nn_approach}
\end{figure}

We present in Figure~\ref{fig:nn_approach} the updated graphics of our approach, highlighting the main components : the distributional part is ensured by a first deep neural network as is each of the mappings.
As each of the networks should be learned, we included the part of the loss functions associated with each network (blue fonts correspond to minimization, whereas red fonts correspond to maximization, see Algorithm section).

\subsection{Training details}
\label{appendix-training_details}

Our experimental evaluations on shape datasets for scalability contrast GW, SGW and HWD. Regarding or classification under isometry transformations, we additionnally consider RI-SGW. Used hyper-parameters for those experiments are detailed below. Notice that SGW, RI-SGW and HWD rely on $K$, the number of projections sampled uniformly over the unit sphere. This $K$ may vary from a method to another. 
\begin{enumerate}
	\item SGW: $K$.
	\item RI-SGW: $\lambda_{\textbf{RI-SGW}}$, the learning rate and $T$, the maximal number of iterations for solving \ref{eq:general_sgw} over the Stiefeld manifold.
	\item HWD: beyond $K$ and the latent space dimension $d$, it requires the parametrization of $\phi$, $\psi$ and $f$ as deep neural networks and their optimizers. For solving the min-max problem by an alternating optimization scheme we use $N$ inner loops and $T$ number of epochs. 
	%  ( (a) optimizing over $f$ with $\psi$ and $\phi$ fixed then (b) optimizing over $\psi$ and $\phi$ with $f$ fixed. 
\end{enumerate}

For SGW and RI-SGW we use the code made available by their authors and cite the related reference \cite{vayer2018fsw} as they require. We use POT toolbox \cite{flamary2021pot} to compute GW distance. 

\paragraph{Scalability} This experiment measures the average running time to compute OT-based distance between two pairs of shapes made of $n$ 3D-vertices. 100 pairs of shapes were considered and $n$ varies in $\left\{100,  250,  500, 1000, 1500, 2000 \right\}$.  

We choose $K_{\text{SGW}}=1000$ (as a default value). 

For HWD, the mapping function $f$ is designed as a deep network with 2 dense hidden layers of size 50. Regarding both $\phi$ and $\psi$, they have also the same architecture as $f$ (with adapted input and output layers) but the hidden layers are 10-dimensional. Adam optimizers with default parameters are selected to train them. Finally we consider $K_{\text{HWD}}=10$, $d=5$, $T=50$, $N=1$ as default values. Notice also that the regularization parameters $\lambda_C$ and $\lambda_a$ are set to 1.

The used ground cost distance for GW distance is the geodesic distance. 

\paragraph{Classification under transformations invariance} For this experiment, we consider the same set of hyper-parameters as for {\bf Scalability} evaluation on shape datasets. Besides, the supplementary competitor RI-SGW was trained by setting $K_{\text{RI-SGW}}=1000 = K_{\text{SGW}}=1000$, $\lambda_{\textbf{RI-SGW}} = 0.01$, $T_{\textbf{RI-SGW}} = 500$. Notice that due to the high-resolution of the meshes (more than 19K three-dimensional vertices), RI-SGW and GW were not able to produce the pairwise-distance matrix used in 1NN classification after several hours.

%GW : geodesic distance
%SGW : number of prof and iteration
%RISGW : number of prof and max_iter
%HWD : all parameters
% \newpage
\section{Additional experimental results}
\label{appendix_more_xp}

\begin{figure}[htbp]
\includegraphics[width=2.5cm]{./figs/3d2d_dsse/gen_000.pdf}
\includegraphics[width=2.5cm]{./figs/3d2d_dsse/gen_999.pdf}
\includegraphics[width=2.5cm]{./figs/3d2d_dsse/gen_1999.pdf}
\includegraphics[width=2.5cm]{./figs/3d2d_dsse/gen_2999.pdf}
\includegraphics[width=2.5cm]{./figs/3d2d_dsse/real.pdf} \\
\includegraphics[width=2.5cm]{./figs/3d2d_sgw/gen_000.pdf}
\includegraphics[width=2.5cm]{./figs/3d2d_sgw/gen_999.pdf}
\includegraphics[width=2.5cm]{./figs/3d2d_sgw/gen_1999.pdf}
\includegraphics[width=2.5cm]{./figs/3d2d_sgw/gen_2999.pdf}
\includegraphics[width=2.5cm]{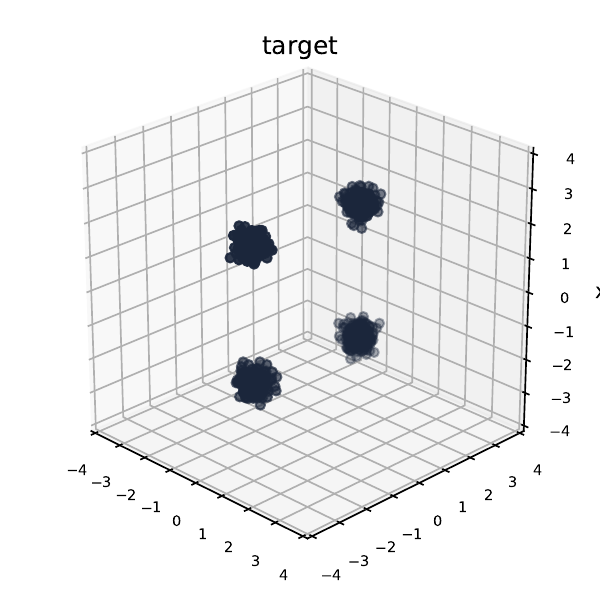} \\
\caption{Comparing (top) HWD and (bottom) SGW on generating 2D distributions from 3D target.}
\label{figure_8}
\end{figure}

\begin{figure}[htbp]
	\includegraphics[width=2.5cm]{./figs/5mode3d_dsse/gen_000.pdf}
	\includegraphics[width=2.5cm]{./figs/5mode3d_dsse/gen_999.pdf}
	\includegraphics[width=2.5cm]{./figs/5mode3d_dsse/gen_1999.pdf}
	\includegraphics[width=2.5cm]{./figs/5mode3d_dsse/gen_2999.pdf}
	\includegraphics[width=2.5cm]{./figs/5mode3d_dsse/real.pdf} \\
	\includegraphics[width=2.5cm]{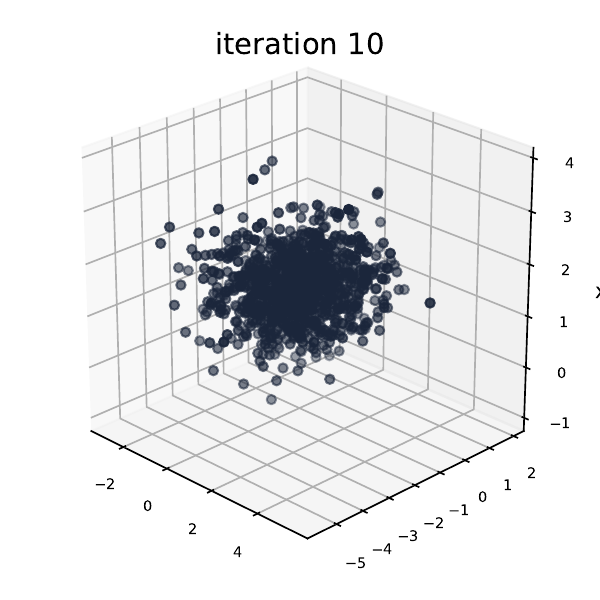}
	\includegraphics[width=2.5cm]{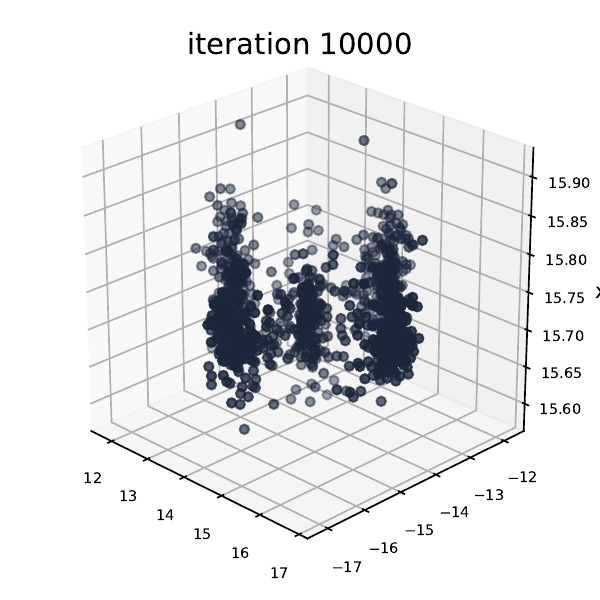}
	\includegraphics[width=2.5cm]{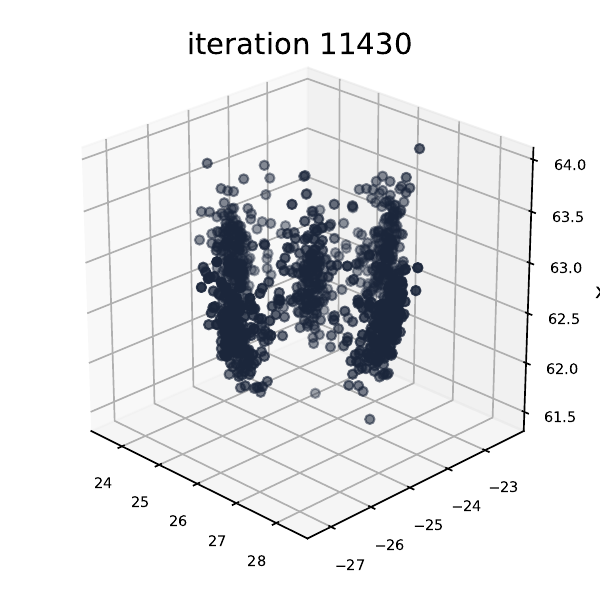}
	\includegraphics[width=2.5cm]{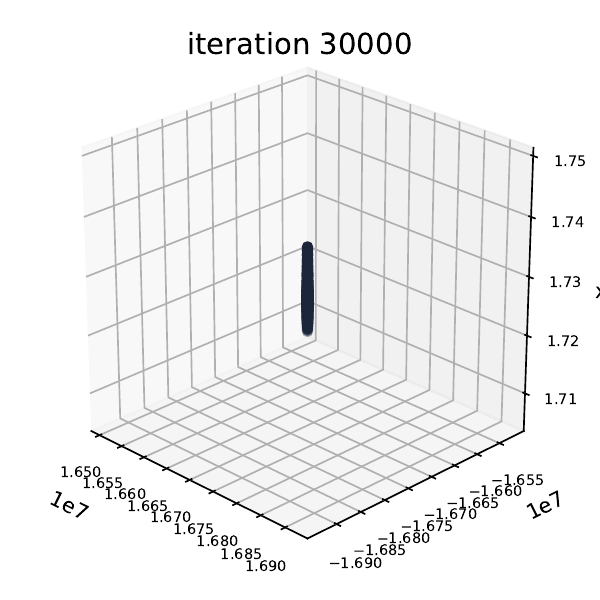}
	\includegraphics[width=2.5cm]{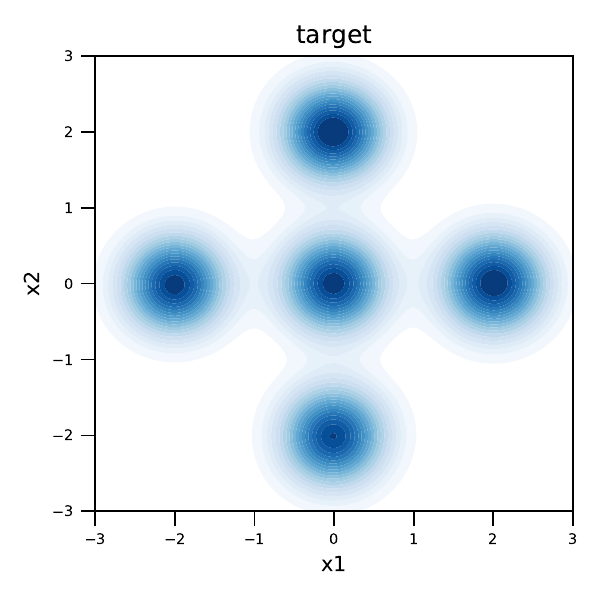} \\
	\caption{Comparing (top) HWD and (bottom) SGW on generating 3D distributions from 2D target.}
    \label{figure_9}
\end{figure}

\begin{figure}[htbp]% [!t]
    \centering
    \includegraphics[width=0.3\linewidth]{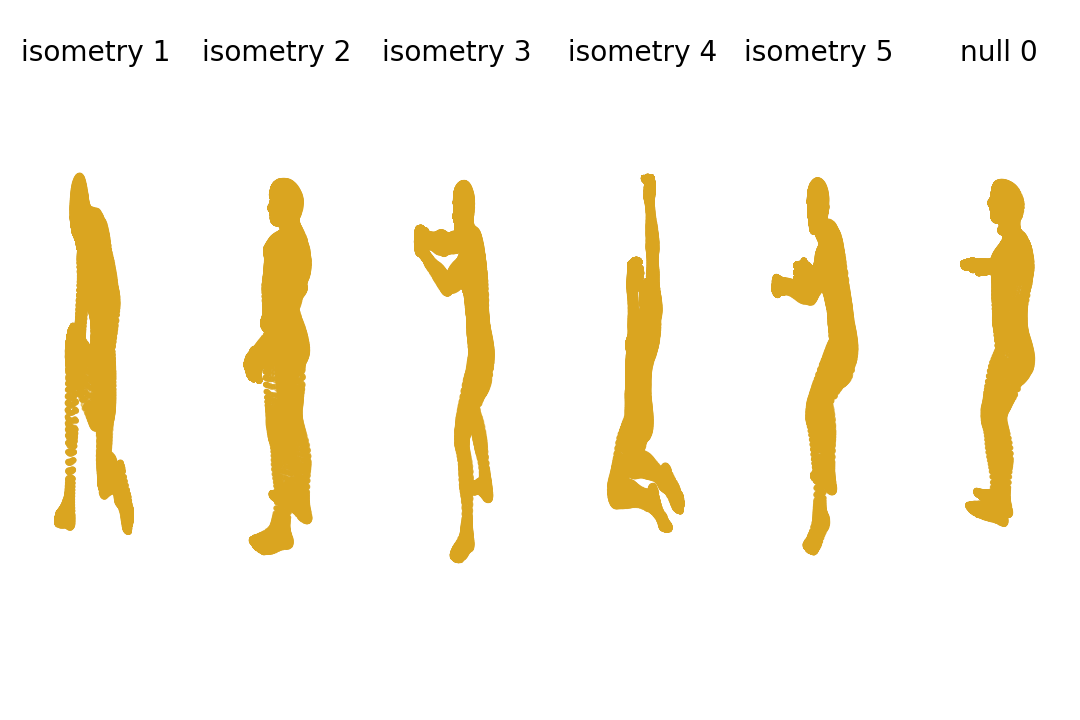} \hspace*{6pt}
    \includegraphics[width=0.3\linewidth]{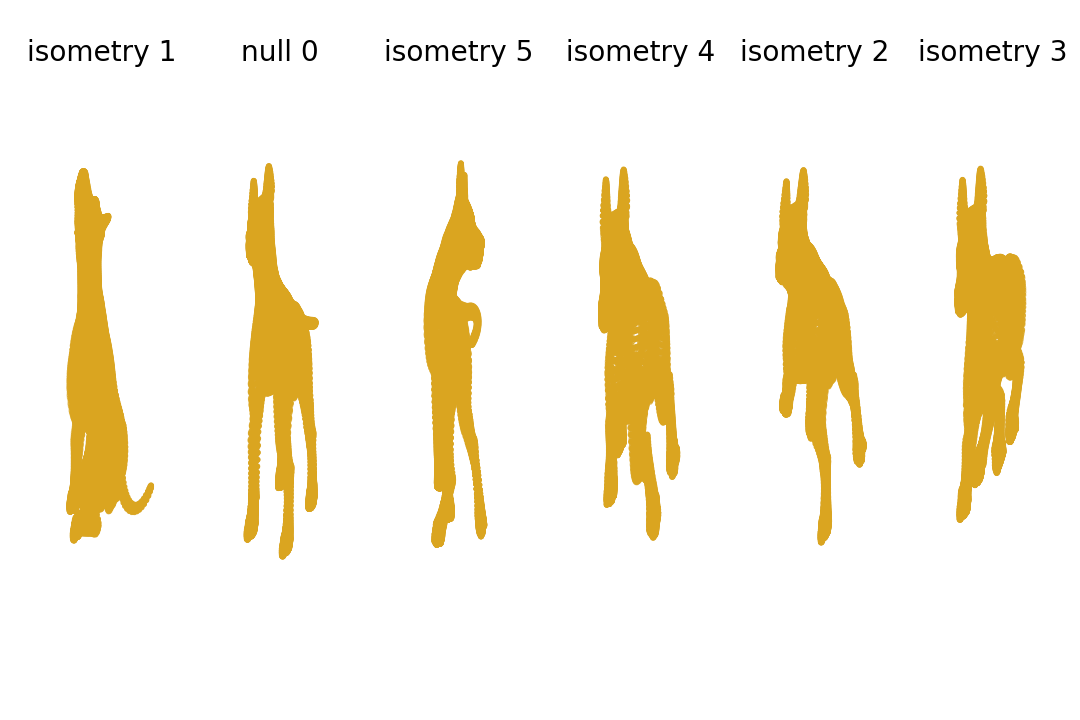}  \hspace*{6pt}
    \includegraphics[width=0.3\linewidth]{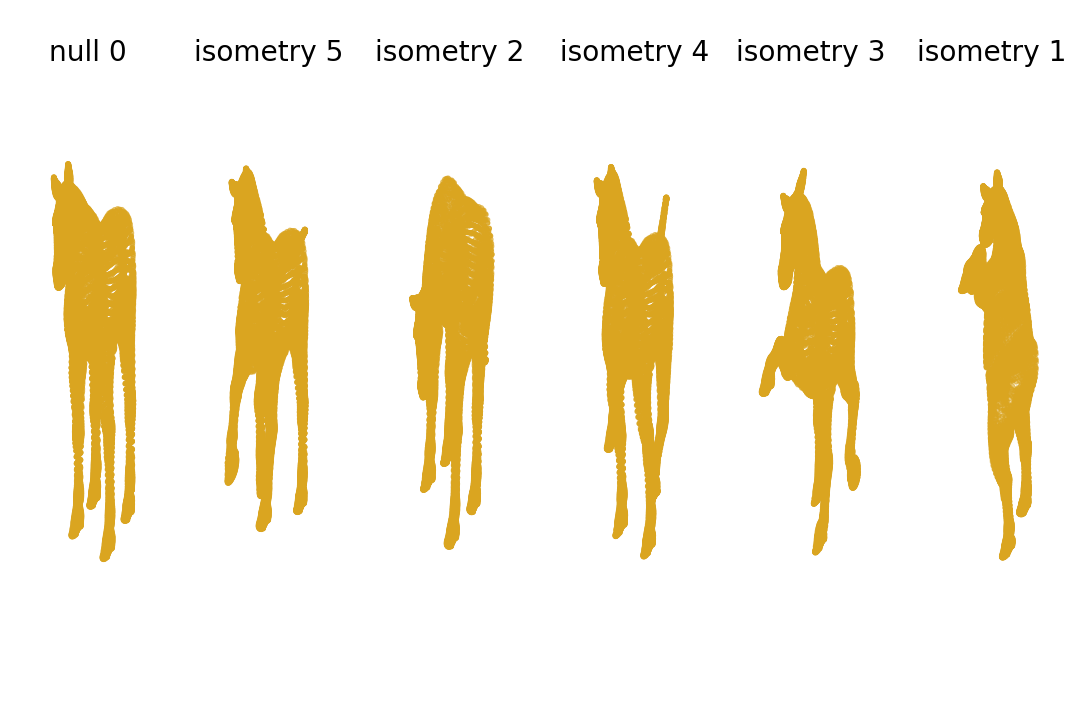}
    \caption{Instances of the shape dataset with null and isometry transformations. The classes are respectively {\tt human,dog} and {\tt horse}. For the experiments of Figure \ref{fig:res_invarianceKNN} we also consider the "topology", "scale", "shotnoise" transformations that respectively amount to deform, to upscale and to add noise to the shapes of each class.}
    \label{fig:robustnessshapes}
\end{figure}

% \section{Broader Impact} \label{sec:broader_impact}
	
% 	Optimal transport tool has proven a significant usefulness for many tasks in machine learning or computer vision especially when dealing with distributions of real entities coming from different spaces. In that setting, this works combines distributional slicing and 1d-Wasserstein distance to propose an approximate discrepancy measure of distributions living in incomparable spaces. This discrepancy share some main features of GW mainly the rotation invariance while exhibiting a favorable computation burden.  We expect the found results will prove useful for efficient computation of Sliced-GW and related applications. As is, we do not foresee any negative societal impacts of the proposed distributional sliced embedding (HWD) OT discrepancy. Rather, we show a computational gain compared with GW distance, and thus, we expect a better carbon-print footprint for applications based on HWD.

\end{document}